\documentclass{article}

\usepackage{adjustbox}
\usepackage{algorithmic}
\usepackage{amsmath,amssymb,amsfonts}
\usepackage{array}
\usepackage{authblk}
\usepackage{bibunits}
\usepackage{booktabs}
\usepackage{caption}
\usepackage{csvsimple}
\usepackage{etoolbox}
\usepackage{graphicx}
\usepackage{hyperref,cleveref}
\usepackage{import}
\usepackage{natbib}[sort]
\usepackage{rotating}
\usepackage[moderate]{savetrees}
\usepackage{siunitx}
\usepackage{subcaption}
\usepackage{textcomp}
\usepackage{xcolor}
\usepackage{lib/alifeconf}
\usepackage{orcidlink}
\usepackage{pdflscape,longtable}
\usepackage{listings}
\usepackage{placeins}
\usepackage{pythonhighlight}

\makeatletter
\let\pragma@iinput=\@iinput
\def\@iinput#1{\xdef\@pragmafile{#1}\pragma@iinput{#1} }
\def\@pragmafile{default}
\def\pragmaonce{%
   \csname pragma@\@pragmafile\endcsname
   \global\expandafter\let \csname pragma@\@pragmafile\endcsname =  
}
\makeatother

\ifdefined\mydraft
\mydraft
\fi

\defaultbibliography{bibl}
\defaultbibliographystyle{apalike-ejor}

\begin{document}

\title{ A Guide to Tracking Phylogenies in Parallel and Distributed Agent-based Evolution Models }

\author[1,2,3,*]{Matthew Andres Moreno\orcidlink{0000-0003-4726-4479}}
\author[4]{Anika Ranjan\orcidlink{0009-0002-8539-8267}}
\author[5,6]{Emily Dolson\orcidlink{0000-0001-8616-4898}}
\author[1,2]{Luis Zaman\orcidlink{0000-0001-6838-7385}}

\affil[1]{Department of Ecology and Evolutionary Biology, University of Michigan, Ann Arbor, United States}
\affil[2]{Center for the Study of Complex Systems, University of Michigan, Ann Arbor, United States}
\affil[3]{Michigan Institute for Data and AI in Society, University of Michigan, Ann Arbor, United States}
\affil[4]{Undergraduate Research Opportunities Program, University of Michigan, Ann Arbor, United States}
\affil[5]{Department of Computer Science and Engineering, Michigan State University, East Lansing, United States}
\affil[6]{Program in Ecology, Evolution, and Behavior, Michigan State University, East Lansing, United States}
\affil[*]{corresponding author: \textit{morenoma@umich.edu}}

\maketitle

\begin{bibunit}

\begin{abstract}
Computer simulations are an important tool for studying the mechanics of biological evolution.
In particular, \textit{in silico} work with agent-based models provides an opportunity to collect high-quality records of ancestry relationships among simulated agents.
Such phylogenies can provide insight into evolutionary dynamics within these simulations.
Existing work generally tracks lineages directly, yielding an exact phylogenetic record of evolutionary history.
However, direct tracking can be inefficient for large-scale, many-processor evolutionary simulations.
An alternate approach to extracting phylogenetic information from simulation that scales more favorably is \textit{post hoc} estimation, akin to how bioinformaticians build phylogenies by assessing genetic similarities between organisms.
Recently introduced ``hereditary stratigraphy'' algorithms provide means for efficient inference of phylogenetic history from non-coding annotations on simulated organisms' genomes.
A number of options exist in configuring hereditary stratigraphy methodology, but no work has yet tested how they impact reconstruction quality.
To address this question, we surveyed reconstruction accuracy under alternate configurations across a matrix of evolutionary conditions varying in selection pressure, spatial structure, and ecological dynamics.
We synthesize results from these experiments to suggest a prescriptive system of best practices for work with hereditary stratigraphy, ultimately guiding researchers in choosing appropriate instrumentation for large-scale simulation studies.
\end{abstract}

\section{Introduction} \label{sec:introduction}

Ongoing advances in computing hardware have potential to unleash transformative, orders-of-magnitude growth in the scale and sophistication of agent-based evolutionary modeling and application-oriented evolutionary computation.
For instance, emerging AI/ML accelerator hardware platforms (e.g., Cerebras Wafer-Scale Engine, Graphcore Intelligence Processing Unit, etc.) currently afford up to hundreds of thousands of processing cores within a single device \citep{lauterbach2021path,jia2019dissecting}.
While these hardware platforms bear limitations characteristic of highly distributed, many-processor computation, their architecture is well-suited for work with agent-based models (ABM), as physical constraints in the layout of processor cores mirror the locally structured interactions typical in ABM.
Nevertheless, significant challenges must be addressed to effectively harness highly distributed, many-processor computation for ABM workloads.

To advance on this front, we propose a fundamental re-frame of simulation that shifts from a paradigm of ``complete,'' deterministic observability of simulation state to instead collect data through dynamic, partial, and potentially best-effort sampling akin to approaches traditionally used to study real-world systems (e.g., ice core samples, paleontological fossils).
The aim of this strategy is to resolve scaling bottlenecks by economizing use of interconnect bandwidth, memory, and disk storage storage and better tolerating intermittent disruption.
Trading a controlled amount of data detail for increased scalability and hardware accelerator compatibility would be highly worthwhile.

Historically, most research using ABM has assumed complete observability of model state.
Indeed, the ability to measure properties \textit{in silico} that would be impossible to observe \textit{in vitro} or \textit{in vivo} is a major benefit of scientific work using ABM.
In the context of evolutionary computation, the existing approach to collecting phylogenetic history is emblematic of this existing complete-observability paradigm.
Typical practice to record phylogeny, the structure of lineage relatedness over evolutionary time, is to accrete every parent-child relationship as it occurs to create a comprehensive tree data structure \citep{moreno2024algorithms}.
This approach produces an exact record and can be highly performant --- particularly when extinct lineages are pruned away \citep{dolson2024phylotrackpy}.

Difficulties arise, however, in extrapolating this approach to a distributed computing context, related to communication overhead of detecting lineage extinctions and sensitivity to data loss \citep{moreno2024algorithms}.
Entirely forfeiting capability to collect phylogenetic information on account of these challenges, though, would significantly reduce the utility of simulation-based evolution experiments and reduce insight into the nuts and bolts of application-oriented evolutionary optimization.
Phylogenetic analysis is integral to much of evolution research, whether conducted \textit{in vivo} or \textit{in silico} \citep{faithConservationEvaluationPhylogenetic1992, STAMATAKIS2005phylogenetics,frenchHostPhylogenyShapes2023,kim2006discovery,lewinsohnStatedependentEvolutionaryModels2023a,lenski2003evolutionary}.
In addition to tracing the history of notable evolutionary events such as extinctions or evolutionary innovations, phylogenetic analysis can also characterize more general questions about the underlying mode and tempo of evolution \citep{moreno2023toward,hernandez2022can,shahbandegan2022untangling,lewinsohnStatedependentEvolutionaryModels2023a}.
One notable application is in evolutionary epidemiology, where phylogenetic structure of pathogens has been used to characterize infection and transmission dynamics within the host population \citep{giardina2017inference,voznica2022deep}.
For application-oriented evolutionary computation, phylogenetic information can even be used to guide evolution toward desired outcomes \citep{lalejini2024phylogeny,lalejini2024runtime,murphy2008simple,burke2003increased}.

Recently developed ``hereditary stratigraphy'' methodology aims to bridge this gap by providing means for extracting phylogenetic information from distributed simulations that are efficient, robust, and straightforward to use \citep{moreno2022hereditary}.
Natural history of biological life operates with no extrinsic provision for interpretable record-keeping, yet phylogenetic analysis of biological organisms has proved immensely fruitful.
Such phylogenetic analyses are possible in biology because mutational drift encodes ancestry information in DNA genomes.
Hereditary stratigraphy methods for decentralized work operate analogously, with ancestry information captured within agent genomes rather than through external tracking.
The idea is to bundle agent genomes with special hereditary stratigraphic annotations in a manner akin to non-coding DNA (entirely neutral with respect to agent traits and fitness) and then use these annotations to perform phylogenetic reconstruction.
The crux of hereditary stratigraphy algorithms, introduced in detail further on, is organization of genetic material to maximize reconstruction quality from a minimal memory footprint \citep{moreno2022hereditary}.

Since it was proposed, experimental work using hereditary stratigraphy has demonstrated viability in extracting information about underlying evolutionary conditions \citep{moreno2024ecology}, even at population scales reaching millions of agents/millions of generations using the 850,000 core Cerebras Wafer-Scale Engine hardware accelerator \citep{moreno2024trackable}.
A number of options exist in configuring hereditary stratigraphy algorithms, but no work has yet systematically investigated how they relate to quality of phylogenetic reconstruction.
In particular, it remains to be established how best to configure hereditary stratigraphy methodology to support use cases varying in scale, memory availability for annotation, and underlying evolutionary conditions.
In this work, we report annotate-and-reconstruct experiments that evaluate reconstruction quality under possible hereditary stratigraphy configurations across a variety of use cases.
We synthesize results from these experiments to suggest a prescriptive system of best practices for work with hereditary stratigraphy.
Analysis covers three primary configurable aspects of hereditary stratigraphy: (1) data structure implementation, (2) temporal data retention policy, and (3) size of stochastic lineage fingerprints.
This work, in conjunction with availability of open-source software library utilities for hereditary stratigraphy \citep{moreno2022hstrat}, is hoped to catalyze means for phylogenetic analysis across a range of large-scale digital evolution projects.

\section{Methods} \label{sec:methods}

The goal of this work is to develop empirically informed best practice recommendations for using hereditary stratigraphy methods to trace ancestry relationships in digital evolution.
This section begins with the conceptual basis of the hereditary stratigraphy approach, covering the checkpoint-based strategy used to assess relatedness.
Two technical aspects of hereditary stratigraphy annotation involved in this work are then introduced in corresponding subsections:
\begin{enumerate}
\item checkpoint retention policy (steady vs. tilted, Section \ref{sec:methods-steady-vs-tilted-algorithms}), and
\item checkpoint storage strategy (column vs. surface, Section \ref{sec:methods-column-vs-surface-algorithms}).
\end{enumerate}
A subsection is provided detailing each of these algorithmic facets.

Attention next turns to experiments conducted to evaluate the reconstruction quality obtained under alternate algorithmic configurations.
For generality, we evaluated reconstruction quality across varied evolutionary scenarios.
Discussion covers the model used to generate reference phylogenies and the set of treatments explored.
Finally, we describe metrics used to measure reconstruction quality, statistical methods, and software used for reconstruction quality experiments.

\subsection{Hereditary Stratigraphy}

Hereditary stratigraphy delivers phylogenetic information analogously to molecular phylogenetics approaches, inferring lineage histories via the tendency for organisms with close hereditary relatedness to exhibit greater sequence similarity \citep{yang2012molecular}.
However, phylogenetic analysis of sequence similarity under a mutational drift model is a nontrivial statistical problem \citep{neyman1971molecular} that can be computationally demanding \citep{konno2022deep,stamatakis2013novel}, data-intensive (e.g., upwards of thousands of base pairs per genome) \citep{parks2009increasing,cloutier2019whole,wortley2005much}, and hindered by challenges arising from back mutation, mutational saturation, selection effects, and branch length differentials \citep{brocchieri2001phylogenetic,moreira2000molecular}.
Although in digital evolution it is possible to use the coding content of model organisms' genomes to estimate relatedness \citep{moreno2021case}, a more general and robust approach is desirable.
To meet this need, hereditary stratigraphy defines a regimen of structured mutation that enables high-quality inference from small amounts of genetic material \citep{moreno2022hereditary}.
The result is a general-purpose framework for phenotypically neutral ``annotations'' that can be affixed to digital organisms' genomes, or even to individual genes, to make their lineages traceable \citep{moreno2022hstrat}.

Hereditary stratigraphy composes annotations as a chronological sequence of checkpoint values.
In each generation, annotations are extended by appending a new random ``fingerprint'' value.
These fingerprints, referred to as ``differentiae'' in the context of hereditary stratigraphy, encode lineage history.
The first pair of mismatching differentiae between two records definitively indicates a split in ancestry.
Conversely, insofar as two annotations share identical differentiae, they likely share common ancestry.
Under this framing, phylogeny reconstruction can be performed agglomeratively using a trie-based procedure.
This approach successively percolates leaf taxa along the tree path of internal nodes consistent with their fingerprint sequence, affixing them where common ancestry ends \citep{moreno2024analysis}.

Because differentiae are randomly generated, it is possible for annotations on diverged lineages to match by chance.
The frequency of such spurious collisions depends on the size of fingerprint values used, e.g., a single bit, a byte, a 32-bit word, etc.
Smaller differentiae reduce annotations' memory footprint but make overestimation of relatedness more likely.
Annotation memory use can also be reduced by discarding old differentiae as generations elapse.
However, sparse retention reduces the number of reference points where divergence between lineages can be detected.
These two mechanisms enable tunable trade-offs between annotation size, inference precision, and inference accuracy.
As such, strategies for differentia sizing and differentia retention are tested in reconstruction quality experiments, described below, to synthesize recommendations for best practice.
The following section delves into the primary dimension of retention policy, steady versus tilted distribution, considered in this work.

\subsection{Steady and Tilted Retention Algorithms}
\label{sec:methods-steady-vs-tilted-algorithms}

\begin{figure}
  \centering
  \includegraphics[width=\linewidth]{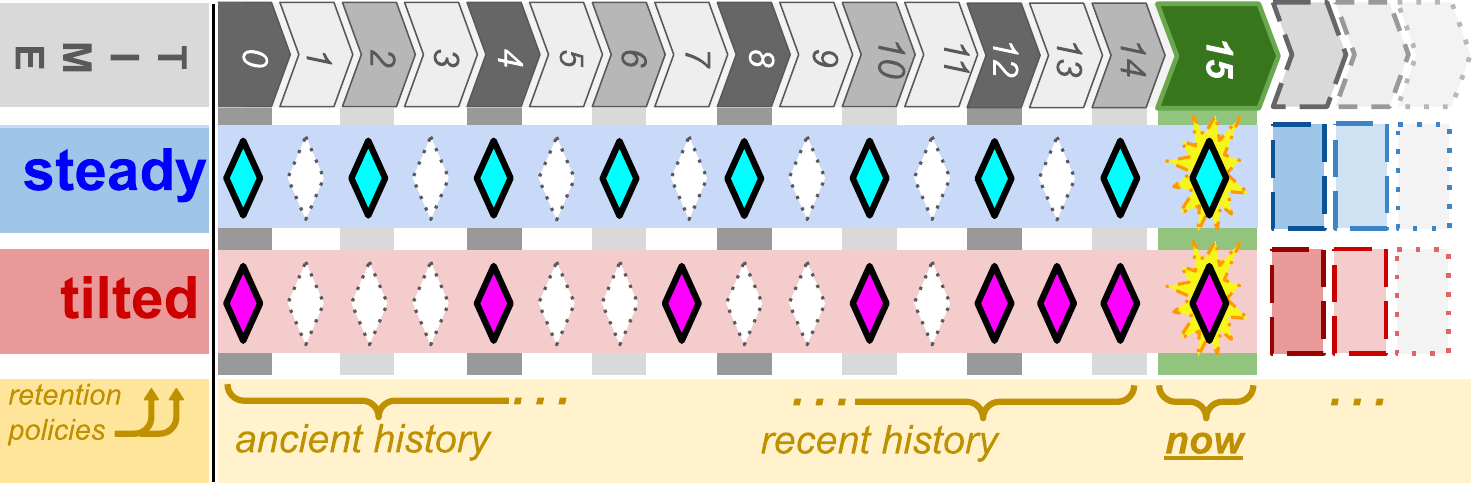}
  \caption{%
  \textbf{Steady versus tilted retention policy.}
  \footnotesize
  \textbf{\textit{Steady policy}} (top) retains differentiae with time points spaced evenly across history.
  \textbf{\textit{Tilted policy}} (bottom) retains differentiae more densely over recent history, giving gap size proportional to time ago.
  Retained differentia are shown as filled diamonds and discarded differentia are shown as empty.
  \textbf{\textit{Hybrid policy}} (not shown) allocates half of available space to hold tilted data and half to hold steady.
  }
  \label{fig:steady-vs-tilted-schematic}
\end{figure}

When pruning differentiae, care must be taken to ensure retention of checkpoint generations that maximize coverage across evolutionary history.
In one possible strategy, retained time points would be spread evenly across history.
We term this strategy ``steady'' \citep{han2005stream,zhao2005generalized}.
Such an approach ensures that last common ancestor (LCA) events can be discerned with consistent absolute precision, no matter when they occurred.

Although the steady approach minimizes worst-case imprecision, there is reason to believe it may fail to allocate precision where it would be most useful for discerning phylogenetic history.
In most evolutionary scenarios, there is a general tendency for phylogenetic events concerning extant taxa to have occurred relatively recently \citep{zhaxybayeva2004cladogenesis}.
A strategy accounting for this fact would retain newer time points at higher density than older time points.
We refer to such an approach, where differentiae are retained in a recency-proportional manner, as ``tilted'' \citep{han2005stream,zhao2005generalized}.

Preliminary experiments have indicated that, in some scenarios, annotations using tilted retention can yield higher quality phylogenetic reconstructions than those using steady retention \citep{moreno2022hereditary}.
Experiments here address this question more thoroughly; we assess the relative performance over a variety of evolutionary scenarios, including those expected to maintain greater amounts of ancient history.
In addition, to assess whether benefits of these two approaches can be combined, some experiments consider a third, ``hybrid'' strategy, where annotations are split half-and-half between steady and tilted strategies.

Figure \ref{fig:steady-vs-tilted-schematic} provides a visual comparison of steady and tilted retention strategies.
The steady strategy retains differentia at regular intervals, while the tilted policy retains differentia in proportion to their recency.
Note that these policies must handle continual accrual of new differentia as generations elapse.
At each step, a new differentia is appended, and --- as required --- old differentia are discarded.
Throughout, constraints on time point coverage and annotation size must be respected.
For more detail on algorithmic aspects of this process, see \citet{moreno2024algorithms,moreno2024structured}.

\subsection{Column and Surface-Based Algorithms}
\label{sec:methods-column-vs-surface-algorithms}

Having just described approaches to deciding which differentiae should be stored, we next consider implementation-level approaches to organize and store them in practice.
A suitable annotation data structure for differentiae curation should:
\begin{enumerate}
\item support efficient update operations (to append new differentia and discard old differentiae),
\item be readily serialized (to exchange annotated genomes between processors), and
\item minimize representational overhead (to reduce annotation memory footprint and inter-process message size).
\end{enumerate}

The last point, minimizing representational overhead, is particularly critical given that typical use calls for single-bit and single-byte differentiae.
Requiring 32- or 64-bit pointers or time point values per differentia would cause bookkeeping overhead to greatly outweigh --- and potentially crowd out --- useful lineage history information.
As such, both approaches considered here --- ``column'' and ``surface''-based storage --- pack differentiae in an array format and rely on positional context to identify them.
For such data to be readily legible, retention policies' curated time points must be directly enumerable \textit{a priori} for any arbitrary generation.

The ``column'' approach arranges differentiae in chronological order, with newest differentiae stored last.
This approach suits use of a dynamic array data structure (e.g., Python \texttt{list}/C++ \texttt{std::vector}), as new additions can be accessioned through an append (e.g., ``\texttt{push\_back}'') operation.
Accordingly, the ``column'' approach supports arbitrary growth in curated collection size.
This property allows for retention policies that provide hard guarantees on inference precision.
\footnote{%
Hard fixed or recency-proportional bounds on differentia gap sizes require orders of growth in retention that are linear and logarithmic, respectively \citep{moreno2024algorithms}.
}
One disadvantage to this approach, though, is that discarding old differentia requires an $\mathcal{O}(n)$ shift-down operation on all subsequent elements.

The ``surface'' approach, in contrast, organizes differentiae directly onto a fixed-length buffer \citep{moreno2024structured}.
Rather than appending as a ``\texttt{push\_back}'' on the array, incoming differentiae are assigned an arbitrary buffer position and directly written there in $\mathcal{O}(1)$ time.
One advantage of this approach is that explicit garbage collection operations are unnecessary; new data simply overwrites that to be discarded.
Another advantage is full use of available space --- after the surface buffer is filled, it is guaranteed that stored differentiae fully utilize available capacity.
Owing to discrepancy between projected upper bounds on retained size and actual usage, this guarantee does not hold for column-based tilted retention where maximum size is capped.
However, to the surface approach's disadvantage, dropping a level of abstraction to operate over buffer sites rather than differentia time points results in less fine-grained control over retention and a somewhat looser adherence to idealized retention patterns.
Additionally, by design, growth beyond the surface's fixed buffer size is not supported.
For a more detailed description of motivation for surface-based algorithms, see \citep{moreno2024trackable}.

In sum, the question of column- versus surface-based algorithms can be characterized as a trade-off between efficiency and exactitude.
Indeed, surface-based algorithms provide order-of-magnitude speedups, as well as better compatibility with low-level, resource-constrained programming environments \citep{moreno2024trackable}.
To assess how, if at all, this trade-off impacts reconstruction quality, we include trials using both approaches in empirical annotate-and-reconstruct experiments, described below.
Note that, in these experiments, we consider only fixed-size annotations.
However, we anticipate nearly all hereditary stratigraphy use cases will apply fixed-size annotation due to benefits of avoiding dynamic memory allocation and variable-length inter-process messaging at runtime.

\subsection{Model System}

This section describes the evolution simulations used to generate reference phylogenies for empirical annotate-and-reconstruct experiments.
To support the large, exact-tracked phylogenies needed for this purpose, we used a simple evolution model.
Genomes comprised a single floating-point value, with higher magnitude corresponding to higher fitness.
We used tournament selection with synchronous generations.
Mutation was applied after selection, with a value drawn from a unit Gaussian distribution added to all genomes.
Experiments used asexual reproduction, with no crossover or recombination.
(Extensions of hereditary stratigraphy to sexual lineages are possible \citep{moreno2024methods}, but not explored in this work.)
Evolutionary runs lasted 100,000 generations.

A key consideration in our experiments is assessing reconstruction quality over a breadth of evolutionary scenarios.
To this end, we applied strong, explicit manipulations of evolutionary conditions to explore hereditary stratigraphy methods across diverse regimes of phylogenetic structure.
One focal variable was phylogenetic diversity, the amount of distinct lineage history maintained within an extant population \citep{tucker2017guide}.
For a fixed population size, phylogenetic richness is increased by coexistence of deep phylogenetic branches.
We explored treatments that promote phylogenetic richness via spatial and ecological structure \citep{moreno2024ecology,gomez2019understanding,valiente2007facilitation}.
Spatial structure in experiments was implemented as a simple island population model and ecological structure was implemented as a simple niche model, respectively.

The island model, used to induce spatial structure, distributed individuals evenly across islands, with selection processes taking place in isolation on each island.
Islands were arranged in a one-dimensional closed ring, and 1\% of population members migrated to a neighboring island each generation.

The niche model, used to induce ecological structure, also applies a simple approach.
Organisms were arbitrarily assigned to a niche at simulation startup, with fixed, equally-portioned population slots assigned to each niche.
In the selection procedure, individuals exclusively compete with members of their own niche.
Every generation, individuals swapped niches with probability $3.05 \times 10^{-8}$ (chosen so one niche swap would be expected every 500 generations at the larger population size and 4,000 generations at the smaller).

We also included a drift treatment, where selection was performed in a fully neutral manner.
Drift conditions also enhance phylogenetic richness, but complement other surveyed treatments in operating through a separate mechanism.

Another objective of employing a highly abstracted model to generate reference phylogenies is generality.
Other work with this simple model suggests that effects on phylogenetic structure induced under treatments surveyed here generalize across model systems, albeit to a less accentuated degree \citep{moreno2024ecology}.
This model system has been established in other existing work, as well \citep{moreno2023toward}.

In addition to structural aspects of phylogeny composition, we also sought to understand the relationship between population scale and reconstruction quality.
Population sizes of both 4,096 ($2^{12}$) and 65,536 ($2^{16}$) were used in experiments.
We anticipate that many use cases of hereditary stratigraphy will collect and analyze only a subset of extant taxa for tractability of data collection, phylogenetic reconstruction, and phylogenetic analysis.
As such, experiments downsampled generated annotations to 500 taxa.
In experiments using larger population size, we also tested reconstructions over larger samples of 8,000 taxa.

\subsection{Experimental Treatments}

For our main experiments, we defined the following ``regimes'' of evolutionary conditions:
\begin{itemize}
  \item \textit{plain}: tournament size 2 with no niching and no islands,
  \item \textit{mild structure}: tournament size 2 with 2 niches and 4 islands,
  \item \textit{rich structure}: tournament size 2 with 8 niches and 64 islands, and
  \item \textit{drift}: tournament size 1 with no niching and no islands.
\end{itemize}



For each evolutionary regime, we tested four arrangements of annotation capacity and differentia size:
\begin{itemize}
  \item 32-bit array,
  \item 64-bit array,
  \item 256-bit array, and
  \item 32-byte array (256-bit size).
\end{itemize}

Differentia size controls the probability of spurious collision, which is $1/2$ for 1-bit differentia and $1/256$ for 1-byte differentiae, while capacity limitations instead affect the time points where divergence between lineages can be compared.

Treatments also considered \textit{steady-versus-tilted} retention policy and \textit{column-versus-surface} implementation.
Recall that the former determines the overall balance of recent-versus-ancient checkpoints kept and the latter determines the implementation-level particulars of discard sequencing.
Experiments using surface implementation, in addition to steady and tilted approaches, also considered a steady/tilted \textit{hybrid} retention policy.

Across all experiments, each treatment comprised 20 replicates.

\subsection{Agglomerative Phylogeny Reconstruction}

To assess how hereditary stratigraphy would reconstruct ground-truth reference trees, we simulated the inheritance of hereditary stratigraphic annotations along reference phylogenies.
This yielded a set of annotations equivalent to what would be attached to extant population members at the end of a run.
Then, we used the agglomerative tree building implementation provided as \texttt{hstrat.build\_tree} in the \textit{hstrat} Python package \citep{moreno2022hstrat}.
Thus, each reconstruction replicate has a directly corresponding reference tree from an exactly tracked evolution run.

Reconstruction was fast, taking less than a second for the 500-leaf tree with the 256-bit annotation and, in optimized mode, about 5 seconds to build the 8,000-leaf tree.
We then apply a postprocessing step, \texttt{hstrat.{\allowbreak}Peel{\allowbreak}Back{\allowbreak}Conjoined{\allowbreak}Leave{\allowbreak}sTrie{\allowbreak}Postprocessor} which accounts for the fact that entirely identical annotations may only arise due to spurious differentia collision because distinct leaf taxa by definition cannot have shared ancestry in their penultimate generation.
This took about 20 seconds for the larger tree.
Additional postprocessing options, including methods to assign timestamps to reconstructed inner nodes, are documented with the library.

The rapid speed of agglomerative reconstructions in this experiment owes, in part, to the synchronous generation structure of reference phylogenies.
Reconstructions over annotations varying in generation count can run slower, owing to complications around differentiae retained by some annotations but already discarded by others.
In other work, we found that the current \textit{hstrat} pure Python \texttt{build\_tree} implementation took about an hour to reconstruct 10,000 tips, a rate slightly faster than 2 nodes per second under optimizations \citep{moreno2024trackable}.
Optimized implementation of \texttt{build\_tree} in a compiled language is on the \textit{hstrat} project roadmap, which will provide for faster reconstructions.
We are also interested in exploring ways to parallelize reconstruction, perhaps by sorting taxa between $n$ subtrees according to their initial differentiae and then filling in those separate trees concurrently.

\subsection{Reconstruction Quality Measures}

\begin{figure}
  \centering
  \includegraphics[width=\linewidth]{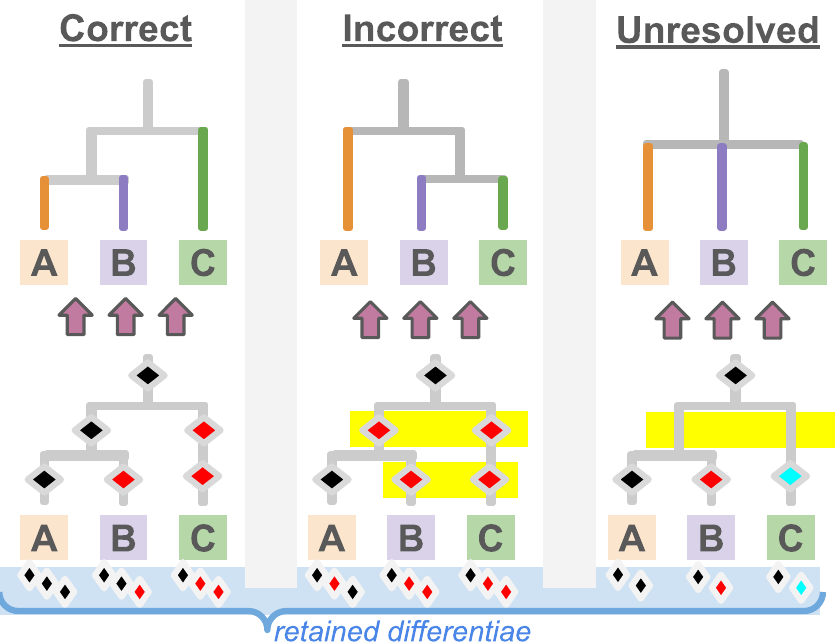}
  \caption{%
    \textbf{Differentia structure and reconstruction outcomes.}
    \footnotesize
    Illustration depicts possible outcomes of reconstruction from hereditary stratigraphy differentia (diamonds) generated and inherited along a two-branch phylogeny (panel bottoms) and resulting reconstruction outcomes (panel tops).
    Diamond placement indicates when differentia were gained and color represents each differentiae's randomly-generated value.
    Diamonds below phylogeny tips summarize inherited hereditary stratigraph record of that taxon.
    \textbf{\textit{Correct reconstruction}} (left panel) occurs when differentia intersperse branching events and differentia value collisions do not occur.
    \textbf{\textit{Incorrect reconstruction}} (center panel) occurs when differentia collisions make unrelated taxa falsely appear related (yellow highlights).
    \textbf{\textit{Unresolved reconstruction}} (i.e., false polytomies; right panel) occurs when differentia do not intersperse branching events but collisions do not occur.
    Note that unresolved reconstructions require differentia size larger than one bit (in order to support $>2$ differentia values), except in the case where more than two differentia records are entirely identical.
  }
  \label{fig:hstrat-failure-modes}
\end{figure}

Assessment of reconstruction quality sought --- in addition to characterizing an overall error measure across hereditary stratigraphy strategies --- to provide diagnostic insight into the nature of reconstruction error produced and the underlying mechanistic reasons it occurs.
Figure \ref{fig:hstrat-failure-modes} depicts two distinct failure modes of hereditary stratigraphy-based reconstruction, which we seek to distinguish.
This example involves three taxa: $A$, $B$, and $C$.
In ground truth, $A$ and $B$ are most closely related and $C$ is an outgroup.
We distinguish three classes of reconstruction outcomes:
\begin{enumerate}
\item \textbf{correct reconstruction}, where retained differentiae suffice to distinguish the $(AB,C)$ branch then the subsequent $(A,B)$ branch;
\item \textbf{incorrect reconstruction}, where spurious differentia collision makes branches appear more closely related than they actually are, e.g., $B$ and $C$ sharing differentiae values by chance, resulting in reconstruction where $B$ and $C$ are inferred as most closely related; and
\item \textbf{unresolved reconstruction}, where differentia necessary to distinguish branching order are not available but subsequent differentia collision does not occur, resulting in artifactual $(A,B,C)$ polytomy.
\end{enumerate}
An exception is the case where several single-bit differentia records are entirely identical, which results in a polytomy where their leaf nodes derive from a common internal node.

We used triplet distance to assess the overall quality of reconstruction \citep{critchlow1996triples}.
This approach considers all possible three-leaf subsets of a phylogeny, and reports the fraction of triplets with topology mismatching the corresponding triplet in a reference tree.
Triplet distance ranges from 0.0 (between identical trees) to 0.5 (between random trees) to a hypothetical maximum of 1.0.
The triplet distance measure requires that phylogenies are rooted \citep{estabrook1985comparison}.
Conveniently, hereditary stratigraphy produces rooted trees, owing to differentia time points being assigned relative to an explicit generation zero.

To discern error arising from unresolved (as opposed to incorrect) reconstruction, we included a second reconstruction quality measure: inner node loss.
This metric quantifies the difference between the number of inner nodes present in the reference tree versus the reconstruction.
It serves as a precision measure, designed to assess the amount of phylogenetic detail lost due to artifactual polytomies.
Inner node loss ranges from 0.0 (for reconstruction with as many inner nodes as reference) to a maximum of 1.0 (reconstruction is a pathological star phylogeny with only one inner node).
Note that a negative inner node loss might be measured if the reconstruction contains more inner nodes than the reference, owing to erroneous overresolution.
This might occur, for instance, due to the inherent inability of bit-width differentia to represent node degrees higher than bifurcation.
For bit-differentia configurations, as remarked above, this inner node loss measure assumes a very specific interpretation: artifactual polytomies occur exclusively when annotation records share all differentiae in common.
In this case, identical annotations are polytomized as leaves of a single inner node.

To further discern error from unresolved versus incorrect reconstruction, we occasionally distinguish an alternate ``lax'' triplet distance from the ``strict'' triplet distance measure described above.
Lax triplet distance differs from strict triplet distance in that it does not penalize triplets that mismatch on account of polytomy.
This measure is useful in isolating incorrect reconstruction from unresolved reconstruction.
However, care should be taken in interpreting lax triplet distance, as the pathological ``star'' tree case where all leaves descend directly from the root in one large polytomy would measure zero reconstruction error under the lax triplet distance measure.
As such, where used, strict triplet distance is also reported.
Where not specified, triplet distance refers to the strict measure.

\begin{figure*}
  \centering

\begin{subfigure}[b]{\textwidth}
\centering
\includegraphics[width=0.5\textwidth]{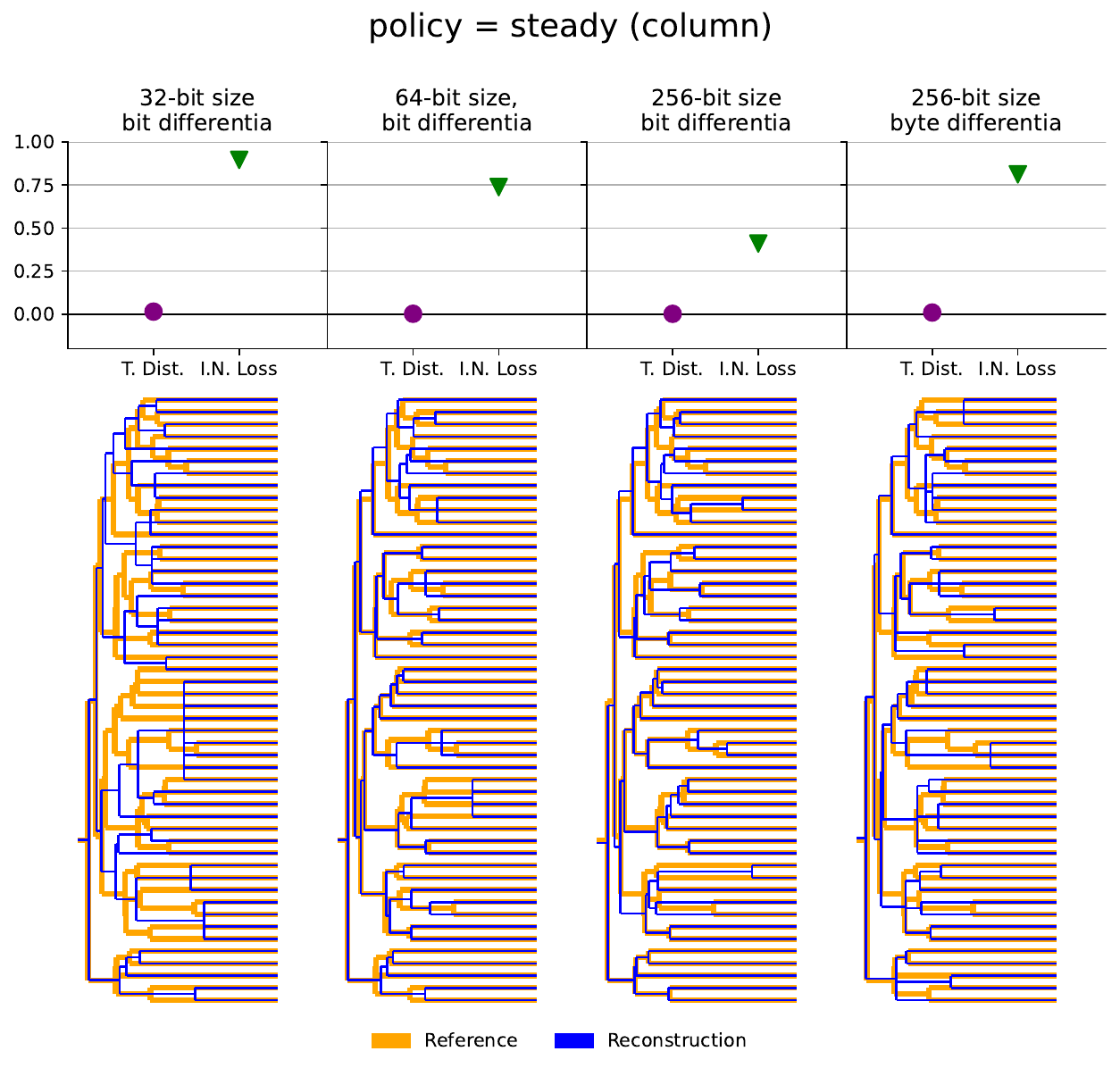}%
\includegraphics[width=0.5\textwidth]{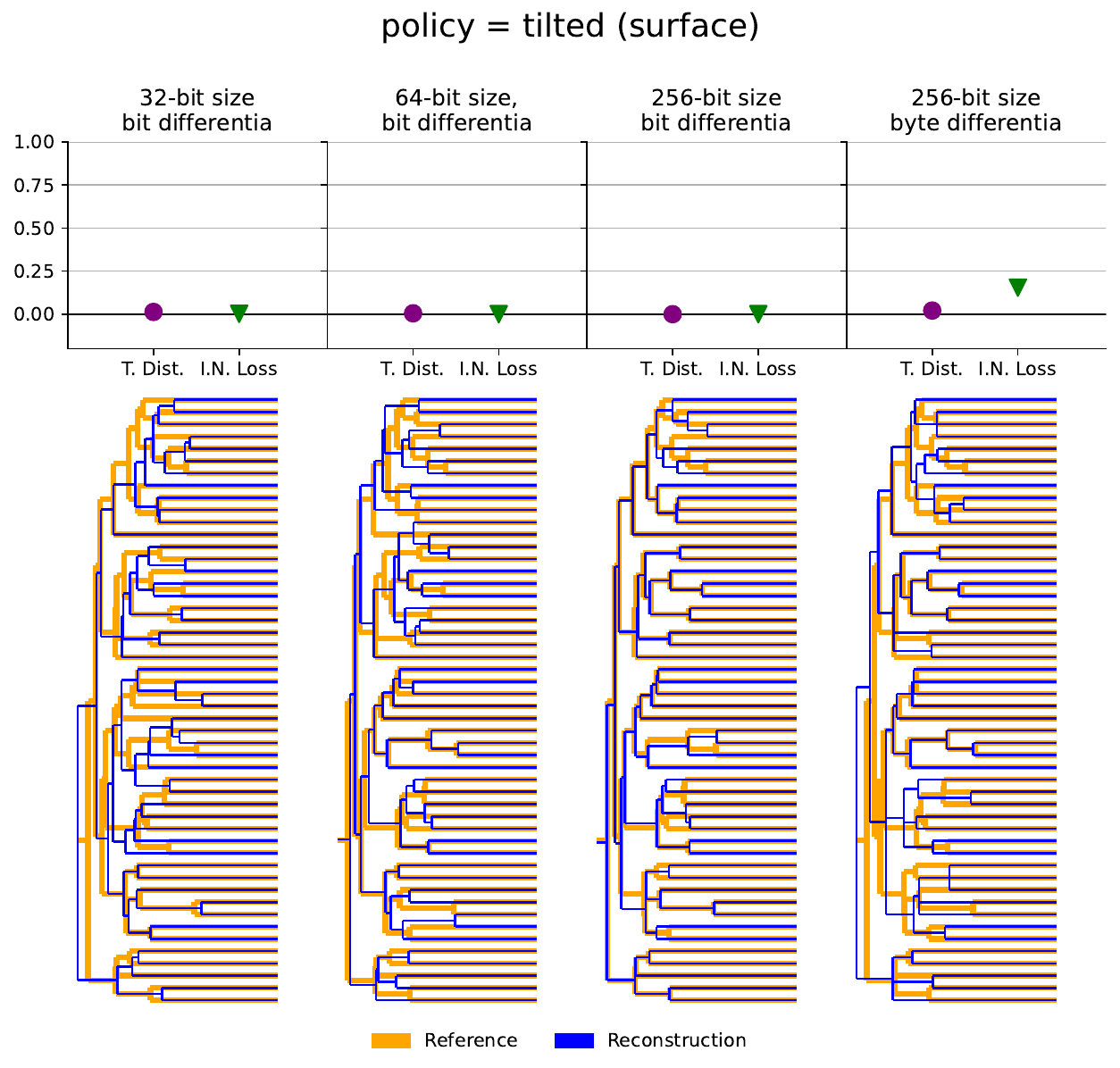}
\caption{\textbf{drift regime} --- high phylogenetic richness}
\label{fig:examplepanel-drift}
\end{subfigure}

\begin{subfigure}[b]{\textwidth}
\centering
\includegraphics[width=0.5\textwidth]{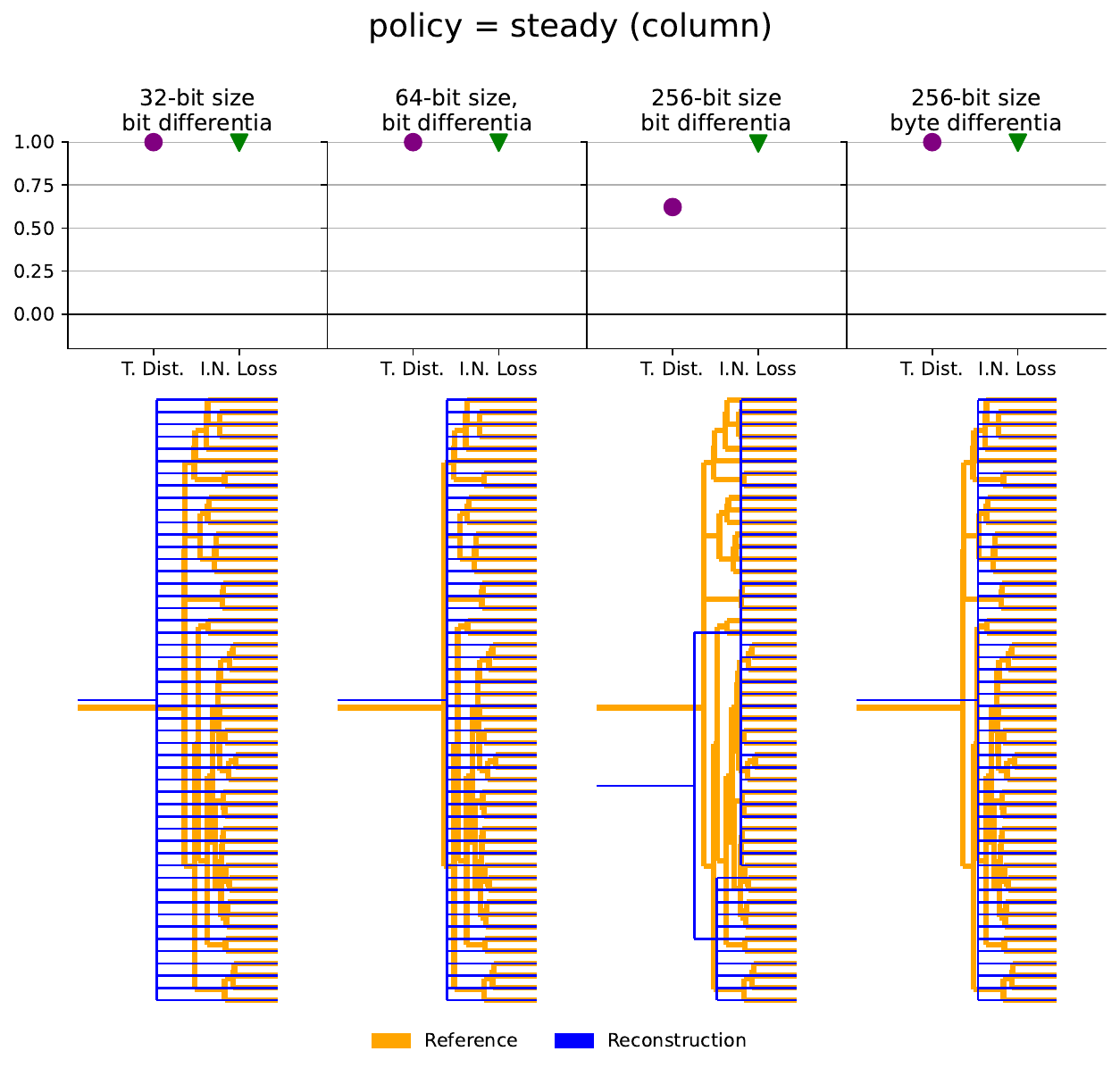}%
\includegraphics[width=0.5\textwidth]{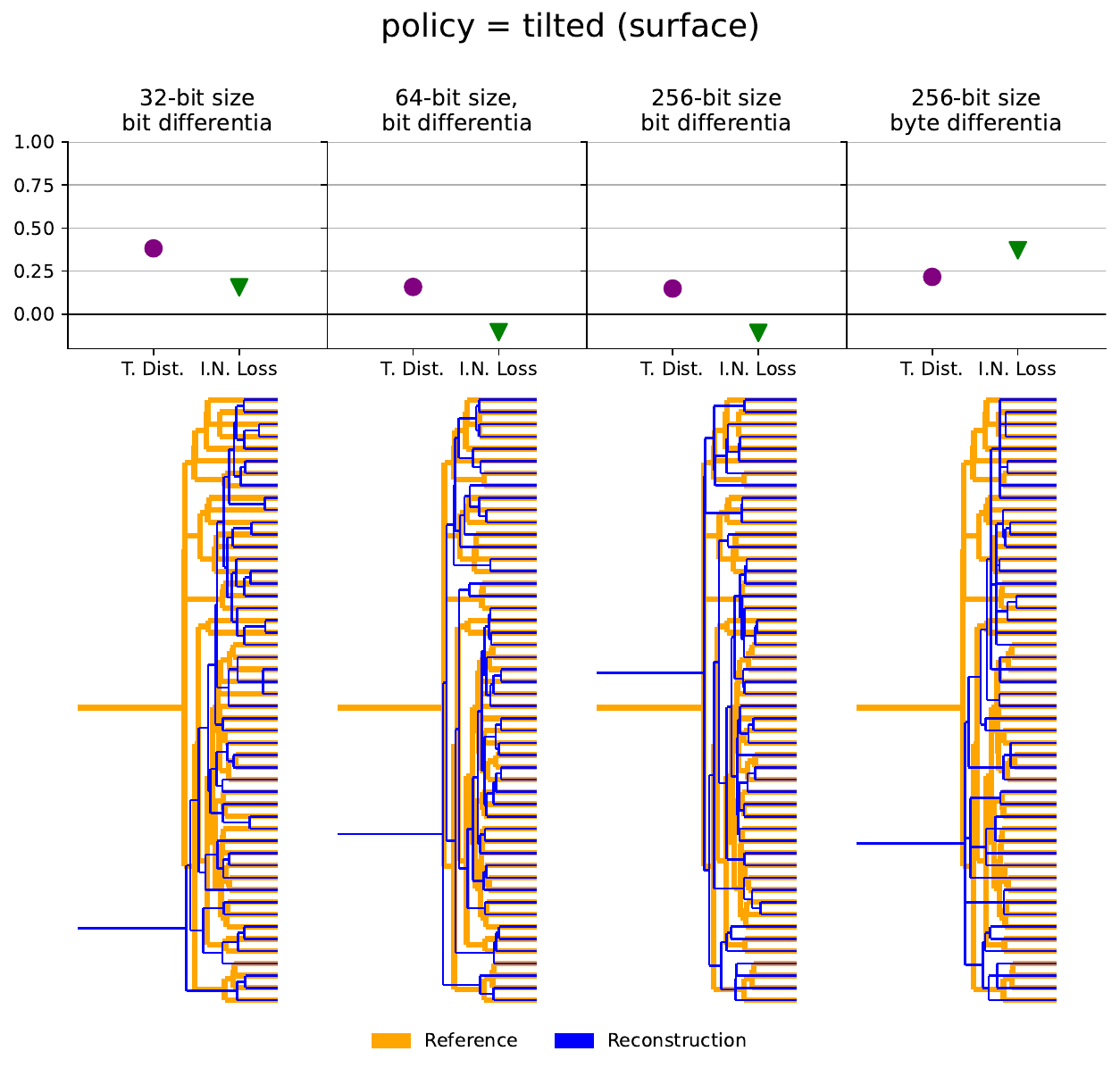}
\caption{\textbf{plain regime} --- low phylogenetic richness}
\label{fig:examplepanel-plain}
\end{subfigure}

  \caption{%
  \textbf{Example Phylogeny Reconstructions and Quality Metric Assessments.}
  \footnotesize
  Comparison of reconstruction to reference tree for steady and tilted policies under drift (\ref{fig:examplepanel-drift}) and plain (\ref{fig:examplepanel-plain}) evolutionary regimes.
  Panel tops show reconstruction quality metrics (triplet distance and inner node loss) and panel bottoms overlay reconstruction (blue) on reference tree (orange).
  Left panels are steady policy and right panels are tilted policy.
  Phylogeny time axes are log scale.
  Note that overlay layout is naive, so can underrepresent agreement between trees; however, comparison is informative to general differences in tree structure.
  Steady policy causes catastrophic comb polytomies in plain regime, where most recent common ancestor among taxa is very recent.
  Steady policy also experiences notable inner node loss under phylogenetically-rich drift scenario, but effect on triplet distance is negligible.
  In all cases, byte differentia configurations have higher, or comparable, triplet distance and inner node loss than correspondingly sized bit differentia configuration.
  }
  \label{fig:examplepanel}

\end{figure*}

Figure \ref{fig:examplepanel} shows example reference phylogenies, corresponding reconstructions, and quality metrics in practice.
The top panel shows trials from the drift treatment, which exhibits high phylogenetic richness, and the bottom panel shows trials from the plain treatment, which exhibits low phylogenetic richness.
(Note that, for legibility, time axes of phylogenies are log-scaled, which somewhat reduces the apparent visual distinction of high-versus-low phylogenetic richness.)
Each panel compares reconstruction results under steady (left) and recency-proportional (right) instrumentation.
Particularly high node loss (green triangle) can be seen under steady retention.
Correspondingly, many large polytomies can be seen in the example steady-retention reconstructions (blue overlaid dendrogram) compared to the corresponding reference phylogeny (orange underlaid dendrogram).
Under the plain treatment, steady retention leads to very high (almost complete) inner node loss, and --- correspondingly --- triplet distance similarity (purple dots) is very poor.
Despite high inner node loss under the drift treatment, triplet distance remains low under steady retention.
Across example cases shown, tilted retention enjoys triplet distance and inner node loss comparable to or better than steady retention.
The main discussion will return to explore this question of steady-versus-tilted retention in greater rigor and depth.

\subsection{Statistical Methods}

Comparisons of reconstruction quality considered both statistical significance (the likelihood an observed difference between treatments might have occurred by chance) and effect size (the magnitude of distinction between treatments relative to outcome variabilities).
We used nonparametric methods for both analyses.
By assessing effect sign, size, and significance across treatment conditions, we described the extent and consistency with which one instrumentation approach outperformed others.

We used Cliff's delta to report effect size.
This statistic describes the proportion of distributional non-overlap between two distributions, ranging from 1 (or -1) if two distributions share no overlap to 0 if they overlap entirely \citep{cliff1993dominance}.
When reporting effect size, we use conventional thresholds of 0.147, 0.33, and 0.474 to distinguish between negligible, small, medium, and large effect sizes \citep{hess2004robust}.
Note that the Cliff's delta statistic tops/bottoms out entirely once two distributions become completely separable.
Where appropriate, we additionally report effects directly in terms of the underlying quality metrics.

We pair effect-size analysis with Mann-Whitney U testing in order to assess evidence that significant differences exist between reconstruction quality under different conditions \citep{mann1947on}.
As our goal was to screen for possible effects, rather than establish the veracity of any one effect, we did not correct for multiple comparisons in assessing statistical significance.
Because incorrectly identifying a lack of difference between conditions would be more harmful than incorrectly identifying the presence of a difference, this approach is conservative in maintaining the sensitivity of our assays.
Where we do detect a significant effect, we additionally report results after applying Bonferroni correction.

When determining the best- and worst-performing among three or more hereditary stratigraphy approaches, we use a nonparametric skimming procedure provided by the \textit{pecking} Python library \citep{moreno2024pecking}.
The procedure first applies a Kruskal-Wallis H-test to determine if there is evidence for significant variation among the sample groups.
If this test fails, then no best- or worst-performing group or groups are identified.
If a significant difference is found, observation ranks are calculated and sample groups are sorted in order of mean rank.
To discern the lowest-ranked group(s), successive Mann-Whitney U-tests are performed between the lowest-rank group and successively higher-ranked groups, adjusting the significance level $\alpha$ for multiple comparisons according to a sequential Holm-Bonferroni procedure.
The overall lowest-ranked group and subsequent groups tested before encountering the first significantly differing group are taken as the best-performing (given that the metric in question is an error measure).
This approach, therefore, identifies the set of lowest-ranked groups that are statistically indistinguishable amongst themselves.
A similar procedure is used to identify a set of highest-ranked groups.

\subsection{Software and Data Availability}

Software, configuration files, and executable notebooks for this work are available via Zenodo at \url{https://doi.org/10.5281/zenodo.11178607}.
Data and supplemental materials are available via the Open Science Framework \url{https://osf.io/n4b2g/} \citep{foster2017open}.

Core hereditary stratigraphy annotation, reference phylogeny generation, and phylogenetic reconstruction tools used in this work are published in the \textit{hstrat} Python package \citep{moreno2022hstrat}.
This project can be visited at \url{https://github.com/mmore500/hstrat}.
On account of recent development of surface-based hereditary stratigraphy algorithms, their source code is currently hosted separately at \url{https://github.com/mmore500/hstrat-surface-concept} \citep{moreno2024hsurf}.
To streamline treatment interoperation, all experiments used underlying \texttt{HereditaryStratigraphicColumn} implementation from \textit{hstrat} and a shim class (available with the surface algorithms) converted the retention patterns that would occur under surface site selection algorithms to column retention policies.
In the medium-term future, we anticipate publishing the surface as a first-class data structure within \textit{hstrat} Python library.

This project uses data formats and tools associated with the ALife Data Standards project \citep{lalejini2019data} and benefited from many pieces of open-source scientific software \citep{sand2014tqdist,2020SciPy-NMeth,harris2020array,reback2020pandas,mckinney-proc-scipy-2010,sukumaran2010dendropy,cock2009biopython,torchiano2016effsize,waskom2021seaborn,hunter2007matplotlib,moreno2024apc,moreno2024qspool,moreno2023teeplot,hagen2021gen3sis,torchiano2016effsize}.

\section{Results and Discussion} \label{sec:results}

This section reports annotate-and-reconstruct experiments comparing phylogeny reconstruction quality obtained across possible hereditary stratigraphy approaches.
These experiments seek to establish a holistic, evidence-driven synopsis of each approach's suitability across experimental use cases.
The following section, ``A Practicioner's Guide to Hereditary Stratigraphy,'' then synthesizes findings to suggest guidelines for selecting appropriate methods to apply in practice.

We delve into three primary aspects of hereditary stratigraphy methodology:
\begin{enumerate}
\item surface- versus column-based implementation,
\item tilted versus steady (versus hybrid) retention, and
\item bit- versus byte-sized differentiae.
\end{enumerate}

In a final set of experiments, we investigate how reconstruction quality fares with increasing phylogeny scale.
Scale-up of subsampled tip count and of underlying population size are both considered.
This question is crucial to application of hereditary stratigraphy for very large simulation use cases --- in assessing the extent, if at all, annotation size would need to be boosted with increased experimental scale.

\subsection{Surface vs. Column Implementation} \label{sec:surface-vs-column}

\begin{figure*}
  \centering
  \begin{subfigure}[b]{0.5\textwidth}
    \centering
    \includegraphics[width=\textwidth]{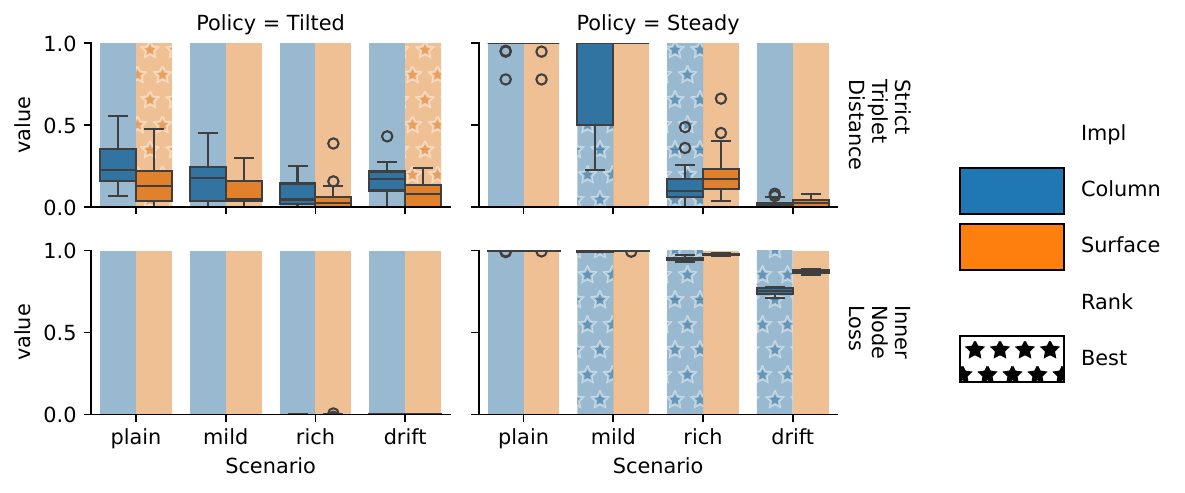}
    \caption{Example reconstruction quality distributions. Lower is better.}
    \label{fig:col-vs-surf-example}
  \end{subfigure}%
  \begin{subfigure}[b]{0.5\textwidth}
    \centering
    \includegraphics[width=\textwidth]{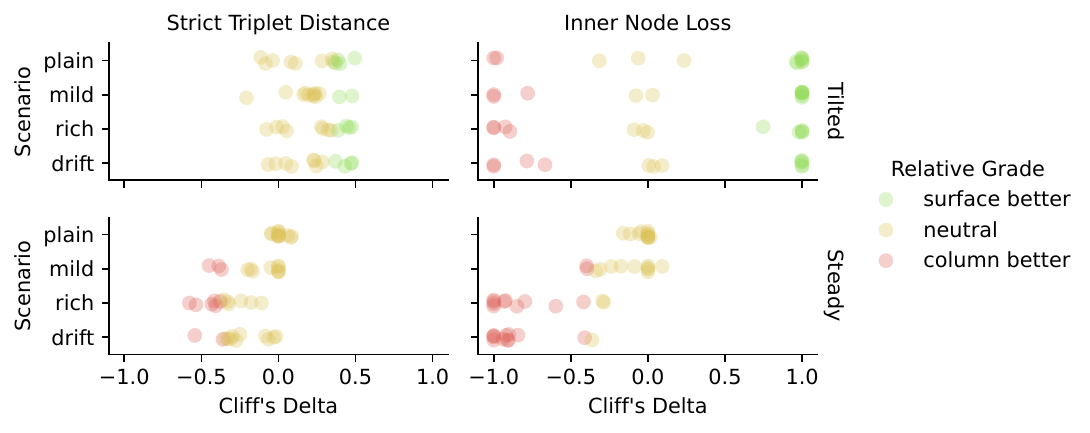}
    \caption{Reconstruction quality comparison outcomes.}
  \label{fig:col-vs-surf-overview}
  \end{subfigure}
  \caption{%
    \textbf{Does column- or surface-based instrumentation give higher-quality reconstruction?}
    \footnotesize
    Subpanel \ref{fig:col-vs-surf-overview} shows effect sizes of column-vs-surface comparisons for triplet distance and inner node loss metrics across sensitivity analysis conditions.
    Color coding indicates a significant outcome (Mann-Whitney U).
    Surface tends to outperform column under tilted policy and vice versa under steady policy.
    Subpanel \ref{fig:col-vs-surf-example} shows reconstruction quality effects for 64-bit size, bit-differentia annotations with population size 65,536, downsample size 500, and 100k generations.
    Background hatching indicates significant outcome.
    See Supplementary Figure \ref{fig:col-vs-surf} for listing of effects by sensitivity analysis condition.
  }
  \label{fig:col-vs-surf-summary}
\end{figure*}

Recall that surface- and column-based annotation implementations differ in how differentiae are organized within a hereditary stratigraphy annotation.
Surface-based implementation takes a lower-level approach that streamlines generation-to-generation updates and ensures full use of available memory space, but sacrifices some control over the temporal distribution of retained differentiae.
Both implementations support tilted and steady retention policies.

Figure \ref{fig:col-vs-surf-summary} compares reconstruction quality for surface algorithms against their corresponding column implementation.
Outcomes differ notably between tilted and steady retention.

\subsubsection{Surface vs. Column for Tilted Retention}

Under tilted retention, triplet distance (a measure of reconstruction accuracy) significantly improves in $14 / 48$ scenarios under surface-based implementation, before correction for multiple comparisons.
Column-based implementation significantly improves reconstruction accuracy in no scenarios when applying the Mann-Whitney U test.
Looking at effect sizes via Cliff's delta, surface-based approaches improve reconstruction quality with at least a small effect size in $33 / 48$ scenarios.
Column-based approaches improve reconstruction quality with a small effect size in 1 /48 scenarios.
Although no Mann-Whitney U tests from any individual scenario are significant after Bonferroni correction, a strong trend exists \textit{across} scenarios of better reconstruction quality from surface-based implementation for tilted retention (exact Binomial test; $p < 0.0001$).

Inner node loss (a measure of reconstruction precision) significantly improves with surface-based implementation in some scenarios ($24 / 48$), and significantly worsens in others ($13 / 48$).
All results remain significant after correction for multiple comparisons.
Notably, shown in Supplementary Figure \ref{fig:col-vs-surf}, inner node loss improvement under surface-based implementation is seen in all treatments with byte-width differentiae, which are more prone to create artifactual unresolved polytomies that drive inner node loss.
In sum for tilted retention, reconstruction quality of surface-based annotations can be considered equivalent or superior across the board to column-based implementation.

\subsubsection{Surface vs. Column for Steady Retention}

In contrast, under steady retention, surface-based implementation achieves worse triplet distance in $11 / 48$ scenarios and better triplet distance in no scenarios.
Only one Mann-Whitney U test remains significant after Bonferroni correction.
However, a significant trend again exists across scenarios, with $25 / 48$ scenarios where surface-based implementation degrades triplet distance by at least a small effect size under Cliff's delta (exact Binomial test; $p < 0.0001$).

Additionally, for steady retention, inner node loss is significantly worse under surface-based in $23 / 48$ scenarios and better in no scenarios.
So, across the board, reconstruction quality of surface-based annotations under steady retention is equivalent or inferior to that of column-based implementation.

\subsubsection{Surface vs. Column --- Synthesis}

Why does the surface-based approach benefit reconstruction under one retention policy but not the other?
Compared to column-based implementation, surface-based implementation allows the tilted algorithm to retain more differentiae within available annotation space.
Whereas the write-only design of the surface-based approach guarantees full use of buffer space, column-based tilted retention wastes some space held in reserve due to algorithmic limitations.
In contrast, column-based steady retention makes full use of available space, giving the surface-based approach no advantage in this regard.
Although both the surface and column approaches prune differentiae through comparable strided decimation procedures, the column implementation more systematically drops decimated differentiae from back to front.
This process ends up preserving more recent differentiae, which --- as we will see in the next set of experiments --- tends to benefit reconstruction quality.

Although surface-based implementation involves a trade-off between performance and reconstruction quality in the case of steady retention, it is notable that surface-based implementation improves both runtime performance and data quality under tilted retention.
Given this result, we next assess the extent to which tilted or steady retention would be preferred in practice across possible evolutionary scenarios.

\subsection{Steady vs. Tilted Retention} \label{sec:steady-vs-tilted}

\begin{figure*}
  \centering
  \begin{subfigure}[b]{0.42\textwidth}
    \centering
    \includegraphics[width=\textwidth]{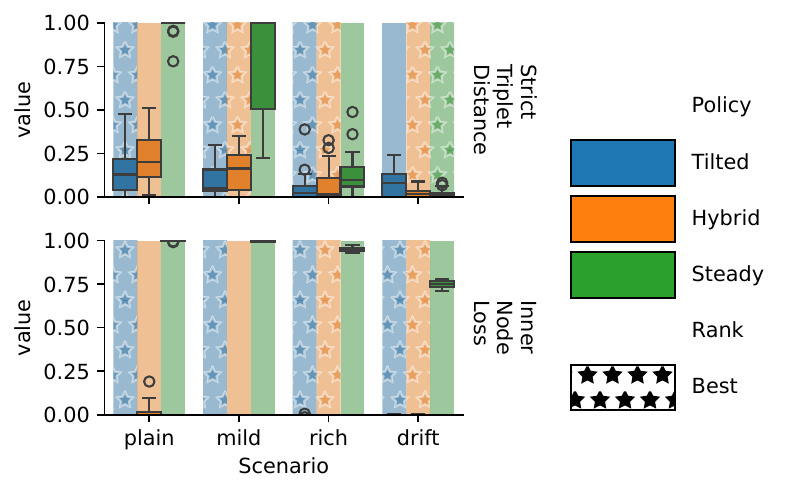}
    \caption{Example reconstruction quality distributions. Lower is better.}
    \label{fig:steady-vs-tilted-summary-example}
  \end{subfigure}%
  \begin{subfigure}[b]{0.58\textwidth}
    \centering
    \includegraphics[width=\textwidth]{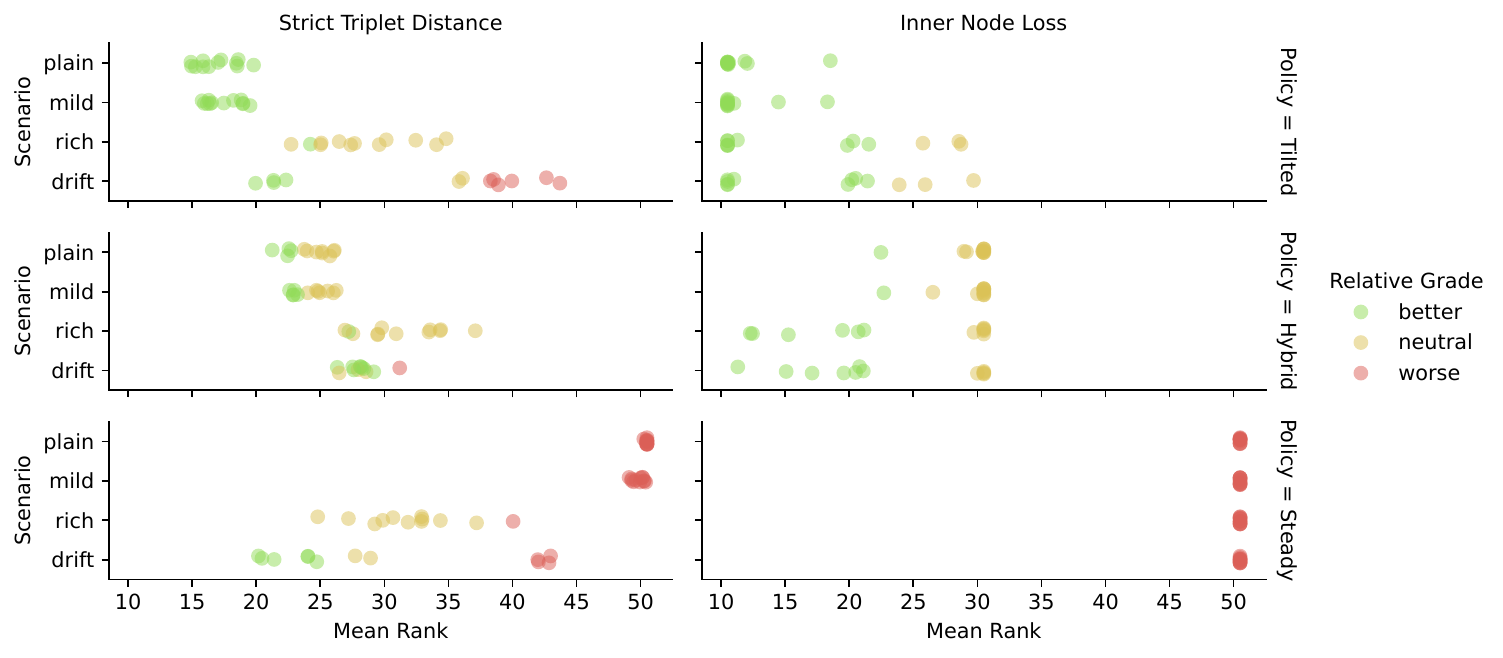}
    \caption{Reconstruction quality comparison outcomes. Lower is better.}
    \label{fig:steady-vs-tilted-summary-overview}
  \end{subfigure}
  \caption{%
    \textbf{How does retention policy affect reconstruction quality?}
    \footnotesize
    Subpanel \ref{fig:steady-vs-tilted-summary-overview} shows mean rank among reconstruction error measures from tilted, hybrid, and steady retention policies across sensitivity analysis conditions.
    Each point represents an independent 20-replicate trial under different evolutionary conditions, instrumentation configuration (e.g., annotation size), and phylogenetic scale (e.g., reconstruction tip count).
    Color coding indicates significant outcome (Kruskal-Wallis H then Mann-Whitney U test).
    Lower is better.
    Tilted policy (top row) performs best in most evolutionary scenarios, except triplet distance under the highly phylogenetically-rich drift regime.
    Steady policy (bottom row) performs worst in most scenarios, except triplet distance under the drift regime.
    Hybrid policy performance has somewhat higher triplet distance reconstruction distance error in the plain and mild scenarios than tilted policy, but is robust to the drift regime.
    Subpanel \ref{fig:steady-vs-tilted-summary-example} shows reconstruction quality effects for 64-bit size, bit-differentia annotations with population size 65,536, downsample size 500, and 100k generations.
    Background hagching indicates significant outcome.
    See Supplementary Figure \ref{fig:steady-vs-tilted} for listing of reconstruction quality outcomes by sensitivity analysis condition.
  }
  \label{fig:steady-vs-tilted-summary}
\end{figure*}

Recall that steady and tilted retention differ in the temporal composition of retained differentiae;
steady policy spaces retained differentia evenly across history, while tilted policy biases toward retaining more recent differentiae.
The question of which policy enables higher quality reconstruction boils down to where precision in discerning the timing of lineage branching events is most useful in resolving evolutionary history.
To ensure even footing, we report results for each retention policy using its best-performing implementation, as established above.
Steady policy uses column implementation and tilted policy uses surface-based implementation.
We also consider a surface-based hybrid policy, which splits buffer space evenly between steady and tilted retention.

\begin{figure*}
  \centering
  \includegraphics[width=\textwidth]{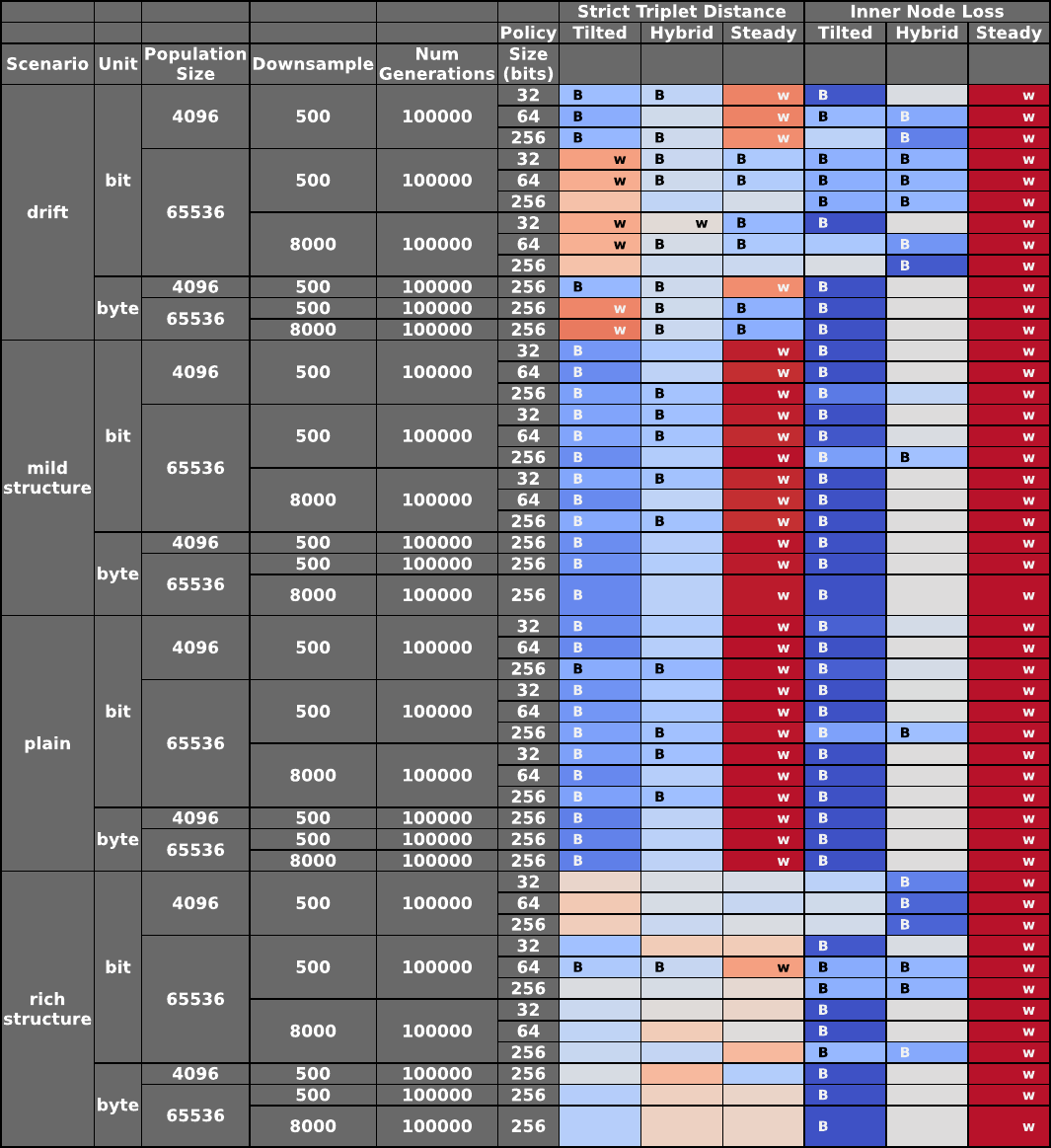}
  \caption{%
    \textbf{Comparison of reconstruction quality across differentia retention strategies.}
    \footnotesize
    Steady retention experiments used column-based implementation; tilted and hybrid retention experiments used surface-based implementation.
    For heatmap charts, B's indicate significantly best and w's significantly worst.
    Heatmap coloring shows nonparametric mean rank among the three algorithms, with blue best and red worst.
  }
  \label{fig:steady-vs-tilted}
\end{figure*}

Figure \ref{fig:steady-vs-tilted-summary} overviews reconstruction quality by retention policy across use case scenarios.
Across the board, steady policy yields phylogenetic reconstruction with heaviest inner node loss.
In contrast, tilted policy exhibits among the lowest inner node loss in all but six surveyed scenarios --- including all byte-differentia trials (Supplementary Figure \ref{fig:steady-vs-tilted}).
The hybrid policy has lower inner node loss in those six scenarios, which all involve evolutionary conditions with high phylogenetic richness.

Steady policy also consistently produces the highest triplet distance error in lower-phylogenetic-richness plain and mild-structure evolutionary scenarios.
However, in scenarios with high phylogenetic richness, triplet distance error under steady retention fares better.
With rich ecological/spatial structure, triplet distance reconstruction error is largely indistinguishable among retention policies.
Further, steady policy triplet error significantly outperforms tilted policy in several instances under drift conditions.
In nearly all of these instances, though, hybrid policy triplet error performs comparably to steady retention.
In low phylogenetic richness plain and mild-structure scenarios, hybrid triplet error is comparable to tilted error in $9 / 24$ scenarios and outperformed by tilted policy in $15 / 24$ scenarios.

Across the inner node loss and triplet error quality measures, tilted retention frequently performs best and steady retention frequently performs worst.
However, tilted retention has worse triplet error in some scenarios with high phylogenetic richness.
Hybrid retention performs more consistently across evolutionary scenarios.
It exhibits consistently intermediate levels of inner node loss that, in absolute terms, tend to be comparable to tilted retention.
Triplet error for hybrid retention is often comparable to the better-performing of steady and tilted retention, or at least intermediate between them.
For greater detail of steady vs. tilted reconstruction quality outcomes broken down by treatment condition, see Supplementary Figure \ref{fig:col-vs-surf}.

\begin{figure*}
  \centering
    \includegraphics[width=\linewidth]{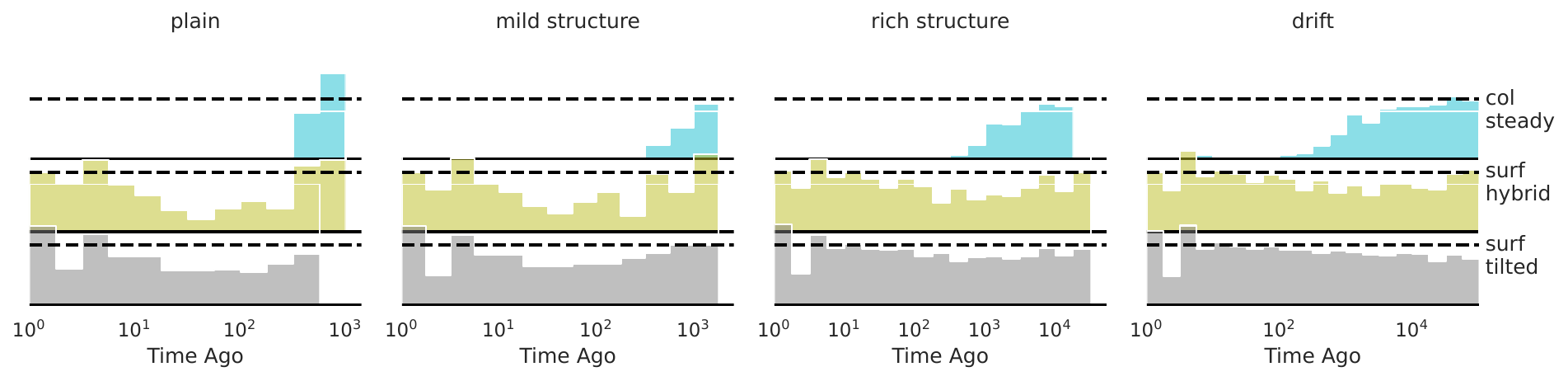}
\caption{%
\textbf{How does retention policy affect the structure of inner node loss?}
\footnotesize
Reconstruction node count densities for 256-bit size, byte differentia treatments.
Histograms depict internal node count relative to reference trees (dashed line), binned by time ago (ranging from most recent to most ancient).
Facet columns differentiate evolutionary regimes and facet rows correspond to steady, hybrid, and tilted retention policies.
Owing to its non-prioritization of retaining recent differentiae, steady policy loses nearly all resolution over recent (c. 100 generations) evolutionary history.
Tilted policy, on the other hand, reconstructs the highest proportion of recent history.
Hybrid policy does not suffer the catastrophic inner node loss over recent events seen under steady policy, but also reconstructs somewhat fewer recent inner nodes compared to pure tilted policy.
}
  \label{fig:recency-structure}

\end{figure*}

Differences in retention of recent differentia explain the substantial advantage of tilted policy in use cases with low phylogenetic richness.
In such scenarios, frequent selective sweeps concentrate phylogenetic history over very recent history, meaning that lineage-branching events giving rise to a contemporary extant population occur over a relatively short period of time.
Discerning these events, therefore, requires densely packed differentia checkpoints over recent history.
Otherwise, in the worst case, taxa would jump from all sharing the same lineage marker at one checkpoint to all having distinct checkpoint markers --- resulting in a catastrophic unresolved polytomy.
Steady retention maintains a uniform gap size between retained differentiae that grows linearly with generations elapsed, meaning that very recent history is thinly covered, if at all.
The consequences of this deficiency can be seen in Figure \ref{fig:recency-structure}, which compares the density of reconstructed nodes to ground truth.
Reconstructions from steady policy are entirely missing branching events over the most recent generations.
Figures \labelcref{fig:bit-vs-byte-summary-byte-outcomes,fig:bit-vs-byte-summary-bit-outcomes} show reconstruction outcomes resulting under this inner node loss.
Under steady policy, very high levels of unresolved reconstruction occur in scenarios with low phylogenetic richness, particularly for recent branching events.
Example reconstructions exhibiting catastrophic comb polytomies characteristic of steady reconstruction of low-richness phylogenies can be seen in \ref{fig:examplepanel}.

\subsection{Bit vs. Byte Differentia Width} \label{sec:bit-vs-byte}

\begin{figure*}
  \centering

\begin{minipage}{\textwidth}
\begin{subfigure}[b]{0.4\textwidth}
\centering
\includegraphics[height=1.2in,trim={0 0 5cm 0},clip]{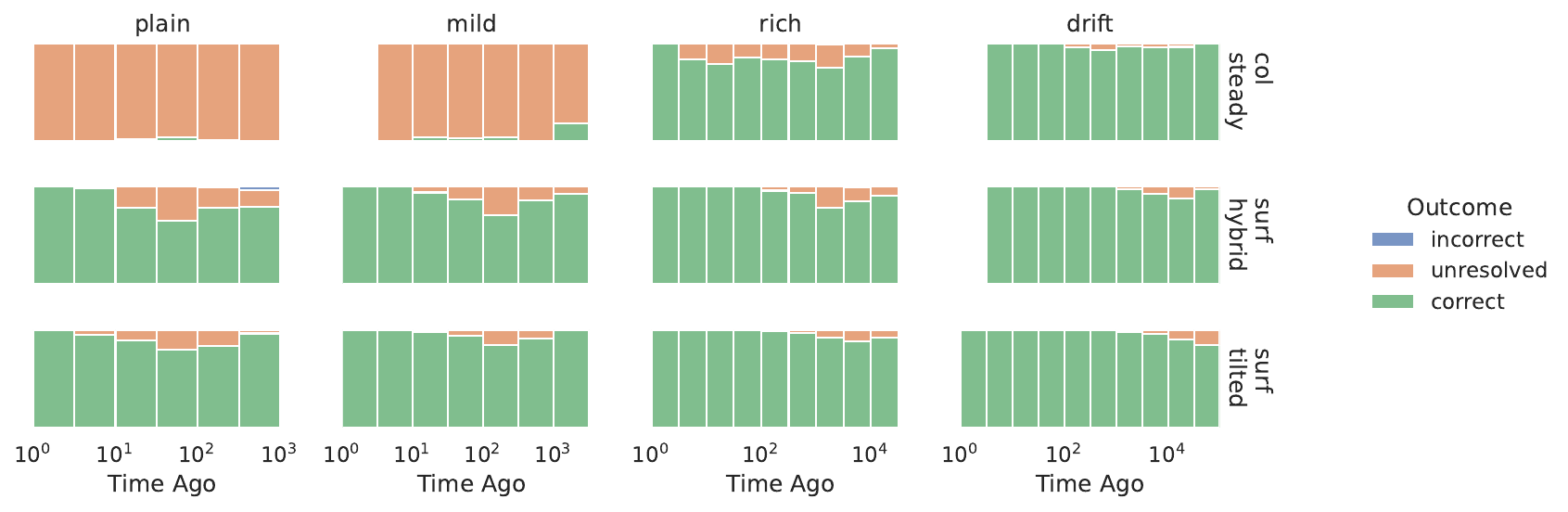}
\footnotesize
\caption{byte differentia outcomes; 256-bit annotation}
  \label{fig:bit-vs-byte-summary-byte-outcomes}
  \end{subfigure}%
\begin{subfigure}[b]{0.6\textwidth}
\centering
\includegraphics[height=1.2in]{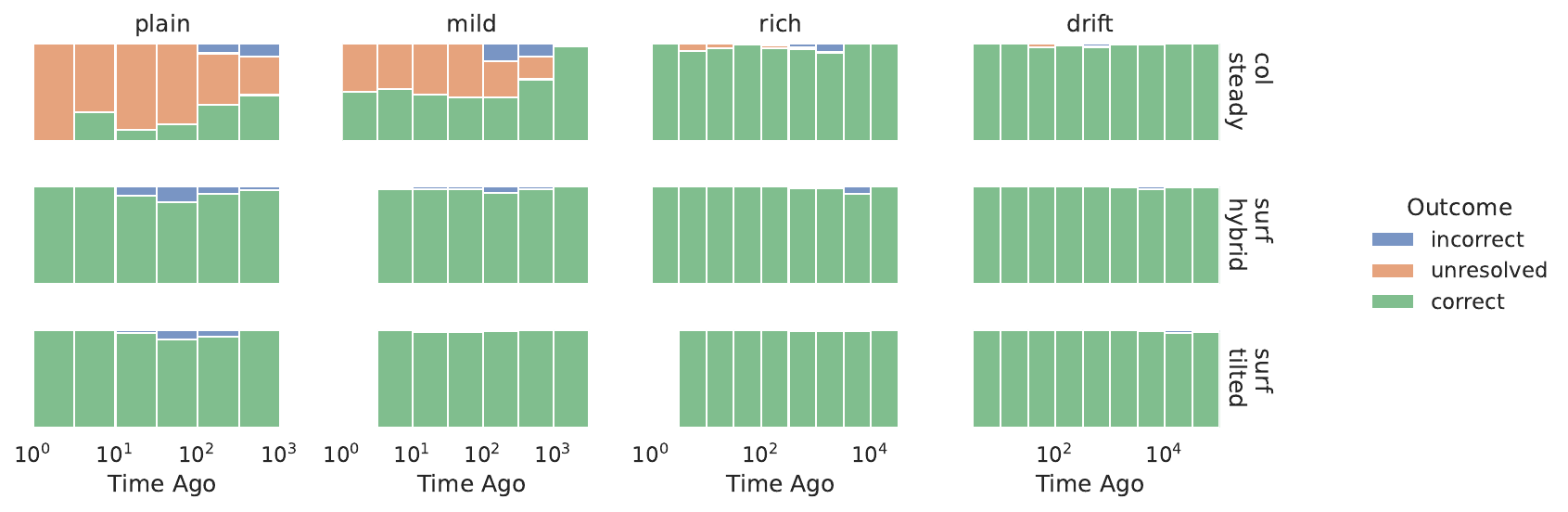}
\footnotesize
\caption{bit differentia outcomes; 256-bit annotation}
\label{fig:bit-vs-byte-summary-bit-outcomes}
\end{subfigure}
\end{minipage}
  \begin{subfigure}[b]{\textwidth}
    \centering
    \includegraphics[width=0.8\linewidth]{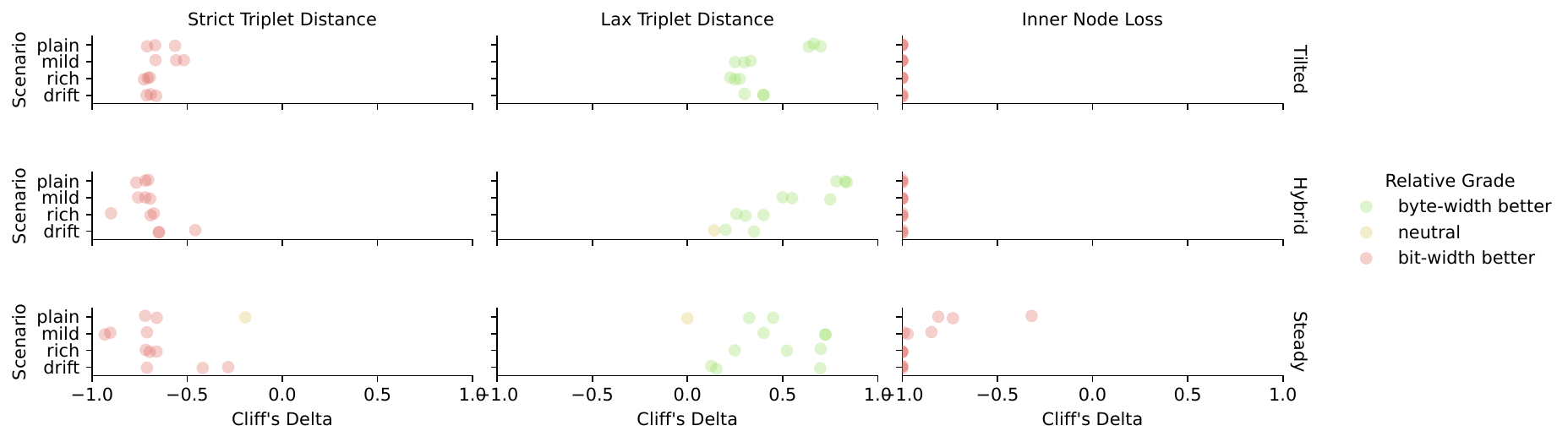}
    \footnotesize
    \caption{Reconstruction quality of byte differentia, relative to bit differentia; 256-bit size annotation}
  \label{fig:bit-vs-byte-summary-quality}
  \end{subfigure}%
\caption{%
  \textbf{How does differentia width affect reconstruction quality?}
  \footnotesize
   Top panels show reconstruction outcomes for phylogenetic branching events (Figure \ref{fig:hstrat-failure-modes}) under bit- and byte-width differentiae, respectively.
   Outcomes are binned by time ago (ranging from most recent to most ancient).
   Bottom panel compares reconstruction quality metrics between reconstructions from annotations with bit- and byte-width differentiae.
   Color coding indicates significance, with red indicating better bit-width differentia performance and green indicating better byte-width differentia performance.
   Byte-width differentiae consistently underperform bit-width differentia in inner node loss and strict triplet distance measures.
   However, byte-width differentiae outperform bit-width differentia in the lax triplet distance measure, which does not penalize polytomy triplets, i.e., isolating incorrect reconstruction from unresolved reconstruction (Figure \ref{fig:hstrat-failure-modes}).
   See Supplementary Figure \ref{fig:bit-vs-byte} for listing of effects by sensitivity analysis condition.
}
  \label{fig:bit-vs-byte-summary}

\end{figure*}

We now consider the role of differentia size in hereditary stratigraphy reconstruction quality.
Intuitively, tuning differentia size would seem to trade-off between accuracy and precision.
Larger differentiae reduce the probability of spurious collision, which falsely makes lineages appear more closely related than they actually are.
Note, in particular, that reconstructions from bit-sized differentia estimate all history as bifurcating, because each checkpoint can only discern two distinct lineages.
On the other hand, for fixed annotation size, widening differentia necessarily reduces differentia count.
Thus, differentia size diminishes the granularity at which branching events can be dated.

\begin{figure*}
  \centering
  \includegraphics[width=\textwidth]{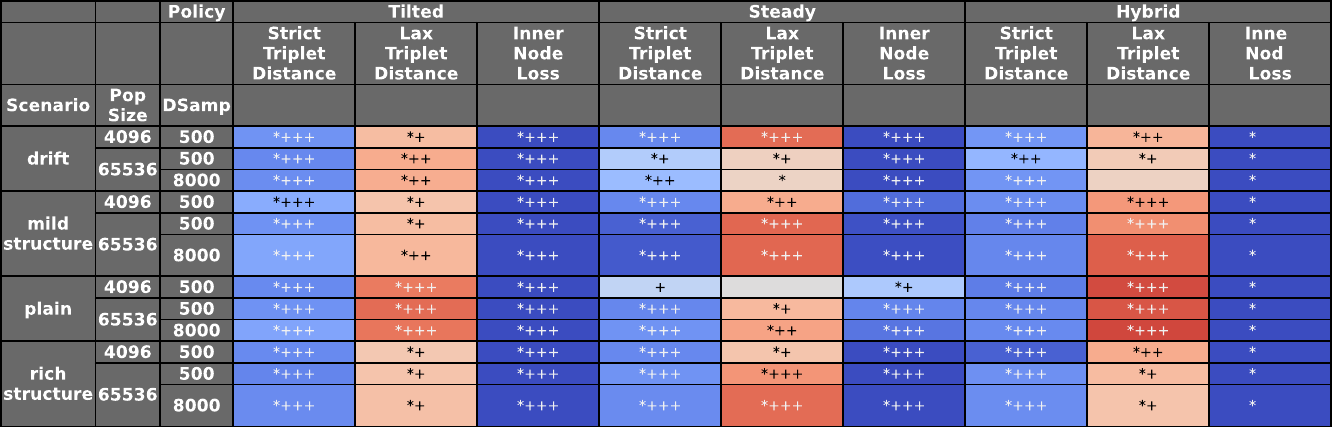}
  \caption{%
   \textbf{Comparison of reconstruction quality from bit- and byte-width differentiae.}
   \footnotesize
    Color coding reflects non-parametric comparison between quality measure values, with red indicating superior byte-width differentia performance and blue indicating superior bit-width differentia performance.
    In cell annotations, +'s indicate small, medium, and large effect sizes using the Cliff's delta statistic and *'s indicate statistical significance at $\alpha = 0.05$ via Mann-Whitney U test.
  }
  \label{fig:bit-vs-byte}
\end{figure*}

To test the effects of differentia width on reconstruction quality, we performed annotate-and-reconstruct experiments across a variety of use case scenarios.
These experiments used 256-bit-sized annotations, comprised of either 256-bit-sized differentiae or 32-byte-sized differentiae.
Figure \ref{fig:bit-vs-byte-summary-quality} overviews the relative performance of bit- and byte-width differentiae on triplet distance and inner node loss quality measures.
(Supplementary Figure \ref{fig:bit-vs-byte} presents these results in greater detail, showing reconstruction outcomes for each treatment condition surveyed.)
However, byte-width differentia consistently produces reconstructions with lower inaccuracy --- as measured by lax triplet distance, which does not penalize unresolved reconstruction.
Byte-width's lower incidence of incorrect reconstruction outcomes is apparent in Figures \labelcref{fig:bit-vs-byte-summary-byte-outcomes,fig:bit-vs-byte-summary-bit-outcomes}, which assess rates of correct, incorrect, and unresolved reconstruction outcomes across evolutionary history.
Unlike bit-width annotations, byte-width methods produce almost no incorrect reconstruction outcomes.
However, owing to unresolved byte-width outcomes, bit-width annotations nonetheless have a higher rate of correct outcomes.
Figure \ref{fig:bit-vs-byte-summary-quality} indeed confirms that, in nearly all cases, bit-width differentiae produce more informative depictions of phylogenetic history, as measured by strict triplet distance.

\subsection{Reconstruction Quality vs. Phylogeny Scale} \label{sec:scaling}
\newcommand{\rulesep}{\unskip\ \vrule\ }

\begin{figure*}
  \centering
  \begin{subfigure}[b]{0.33\textwidth}
    \centering
    \includegraphics[height=1.7in,trim={0 0 5cm 0},clip]{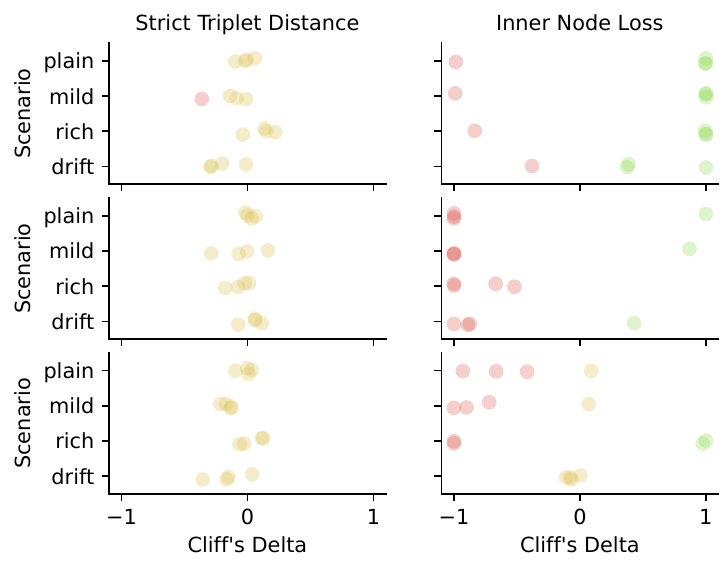}
    \caption{tilted retention policy (surface)}
  \end{subfigure}%
  \rulesep %
  \begin{subfigure}[b]{0.26\textwidth}
    \centering
    \includegraphics[height=1.7in,trim={2cm 0 5cm 0},clip]{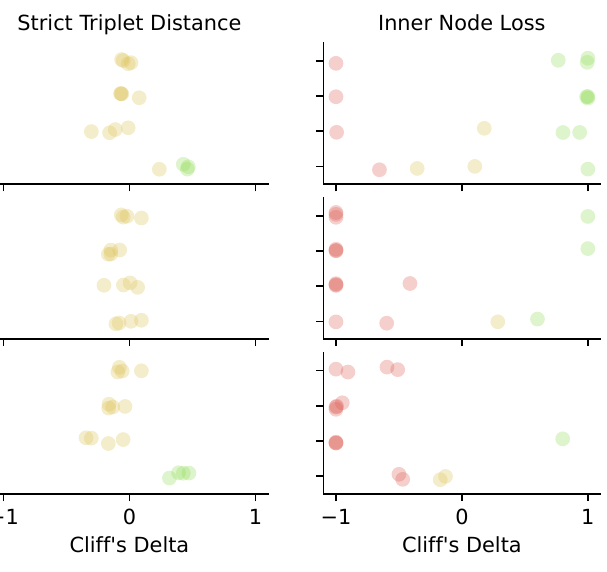}
    \caption{hybrid retention policy (surface)}
  \end{subfigure}%
  \rulesep %
  \begin{subfigure}[b]{0.39\textwidth}
    \centering
    \includegraphics[height=1.7in,trim={2cm 0 0 0},clip]{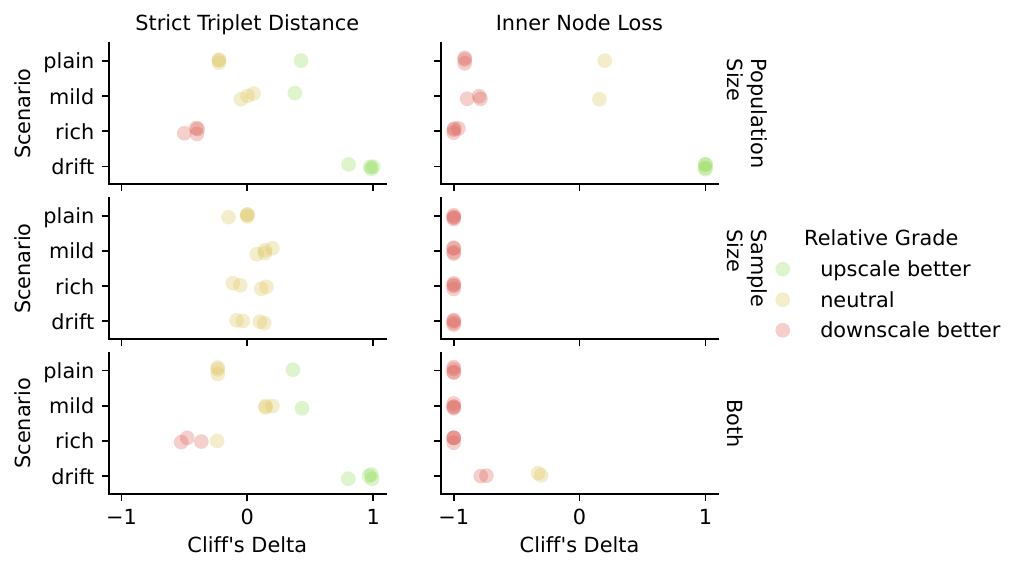}
    \caption{steady retention policy (column)}
  \end{subfigure}
  \caption{%
  \textbf{How does scale impact reconstruction quality?}
  \footnotesize
  Each dot shows a Cliff's Delta effect size of downscale-vs-upscale comparison under one sensitivity analysis condition.
  Color coding indicates a significant outcome (Mann-Whitney U).
  In order, rows show scaling outcomes for increasing population size, population subsample size (i.e., tree tip count), and both these factors.
  For tilted and hybrid policies, scaling had minimal effects on triplet distance and no systematic effect on inner node loss.
  Inner node loss worsened when scaling sample size, with and without scaling population size.
  Scaling effects were more variable under steady retention policy.
  See Supplementary Figures \labelcref{fig:dsamp-popsize-scale-hybrid,fig:dsamp-popsize-scale-steady,fig:dsamp-popsize-scale-tilted} for listing of effects by sensitivity analysis condition.
  }
  \label{fig:scaling-summary}
\end{figure*}

\begin{figure*}
  \centering
  \includegraphics[width=\textwidth]{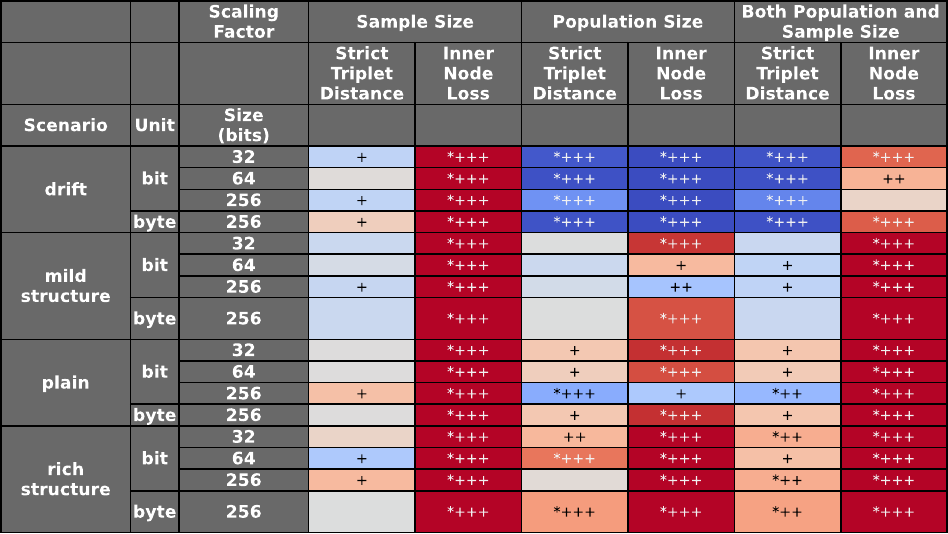}
  \caption{%
   \textbf{Comparison of reconstruction quality between small and large downsample and population sizes under steady retention policy.}
   \footnotesize
    First column considers sample size in isolation, second column considers scaling population size in isolation, and third column considers scaling population and sample size together.
    Color coding reflects non-parametric comparison between reconstruction quality measure values, with red indicating degraded reconstruction quality at larger scale and blue indicating improved reconstruction quality at larger scale.
    Larger downsample size is 8,000 taxa and smaller downsample size is 500 taxa.
    Larger population size is 65,536 and smaller population size is 4,096.
    Experiments used steady retention policy with column-based implementation.
    In cell annotations, +'s indicate small, medium, and large effect sizes using the Cliff's delta statistic and *'s indicate statistical significance at $\alpha = 0.05$ via Mann-Whitney U test.
  }
  \label{fig:dsamp-popsize-scale-steady}
\end{figure*}

\begin{figure*}
  \centering
  \includegraphics[width=\textwidth]{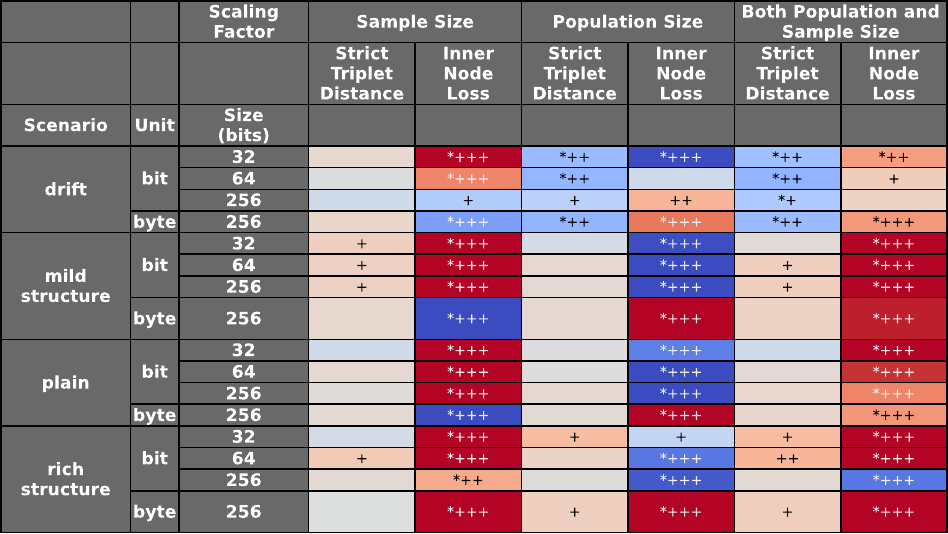}
  \caption{%
   \textbf{Comparison of reconstruction quality between small and large downsample and population sizes under hybrid retention policy.}
   \footnotesize
    First column considers sample size in isolation, second column considers scaling population size in isolation, and third column considers scaling population and sample size together.
    Color coding reflects non-parametric comparison between reconstruction quality measure values, with red indicating degraded reconstruction quality at larger scale and blue indicating improved reconstruction quality at larger scale.
    Larger downsample size is 8,000 taxa and smaller downsample size is 500 taxa.
    Larger population size is 65,536 and smaller population size is 4,096.
    Experiments used hybrid retention policy with surface-based implementation.
    In cell annotations, +'s indicate small, medium, and large effect sizes using the Cliff's delta statistic and *'s indicate statistical significance at $\alpha = 0.05$ via Mann-Whitney U test.
  }
  \label{fig:dsamp-popsize-scale-hybrid}
\end{figure*}

\begin{figure*}
  \centering
  \includegraphics[width=\textwidth]{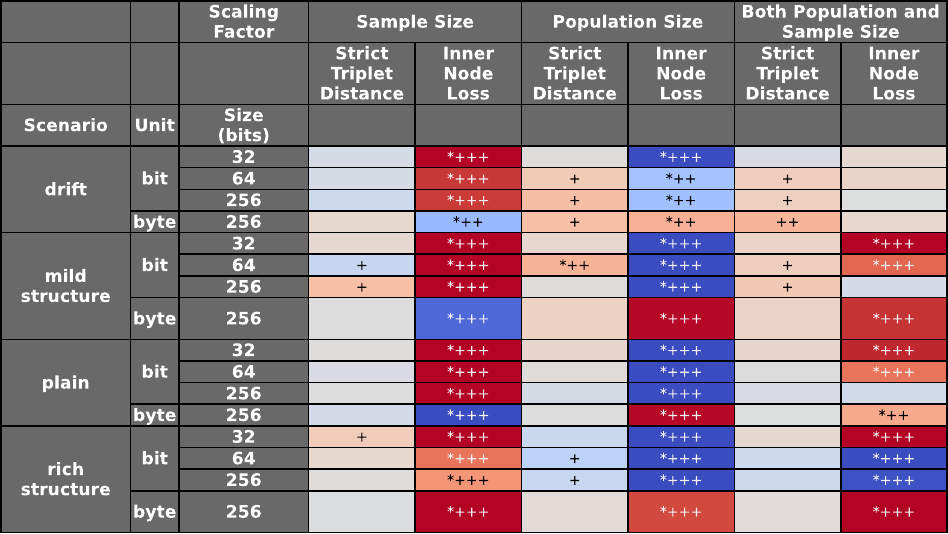}
  \caption{%
   \textbf{Comparison of reconstruction quality between small and large downsample and population sizes under tilted retention policy.}
   \footnotesize
    First column considers sample size in isolation, second column considers scaling population size in isolation, and third column considers scaling population and sample size together.
    Color coding reflects non-parametric comparison between reconstruction quality measure values, with red indicating degraded reconstruction quality at larger scale and blue indicating improved reconstruction quality at larger scale.
    Larger downsample size is 8,000 taxa and smaller downsample size is 500 taxa.
    Larger population size is 65,536 and smaller population size is 4,096.
    Experiments used tilted retention policy with surface-based implementation.
    In cell annotations, +'s indicate small, medium, and large effect sizes using the Cliff's delta statistic and *'s indicate statistical significance at $\alpha = 0.05$ via Mann-Whitney U test.
  }
  \label{fig:dsamp-popsize-scale-tilted}
\end{figure*}

The goal of hereditary stratigraphy methods is to empower new frontiers in digital evolution that harness parallel and distributed computing methods to realize scale-dependent experiments tackling phenomena like evolutionary transitions, eco-evolutionary dynamics, and open-endedness \citep{moreno2022exploring,dolson2021digital,channon2019maximum}.
We also hope it will prove useful in future application-oriented work using massively parallel and distributed computing for evolutionary optimization.
Given these objectives, the scaling behavior of hereditary stratigraphy is of key concern.
Here, we consider two aspects of potential scale-up: (1) the number of taxa sampled for phylogenetic reconstruction and (2) the size of the underlying population.
Experiments test how reconstruction quality fares with changes in scale along both fronts, across a variety of use case scenarios.
Supplementary Figures \labelcref{fig:dsamp-popsize-scale-hybrid,fig:dsamp-popsize-scale-steady,fig:dsamp-popsize-scale-tilted} detail the use case scenarios surveyed and summarize reconstruction quality outcomes of phylogeny scale-up under each scenario.

Scaling results for reconstruction accuracy, measured by triplet distance, are promising.
Under tilted and hybrid policies, triplet distance error is robust across kinds of phylogenetic scaling --- population size, sample size, and both simultaneously.
Reconstruction accuracy decreases significantly in only one case under the tilted policy.
Results under the steady policy are more variable, with triplet distance worsening in 7 cases but improving in 11 cases.

Inner node loss, under hybrid and tilted retention, is generally also robust to population scaling.
For these policies, inner node loss worsens only in use cases with byte-width differentiae (Supplementary Figures \labelcref{fig:dsamp-popsize-scale-hybrid,fig:dsamp-popsize-scale-tilted}).
In other cases, inner node loss improves with population scale.
Sample size scaling, however, tends to aggravate inner node loss --- whether in isolation or in concert with population size scaling.
As an exception, we do not see inner node loss worsen with larger sample size in some cases with byte-width differentia under tilted and hybrid policies (Supplementary Figures \labelcref{fig:dsamp-popsize-scale-hybrid,fig:dsamp-popsize-scale-tilted}).
Likewise, inner node loss remains stable for tilted retention under drift conditions (Supplementary Figure \labelcref{fig:dsamp-popsize-scale-tilted}).

\section{A Practitioner's Guide to Hereditary Stratigraphy} \label{sec:synthesis}

This section shifts focus from an analytical, hypothesis-driven framing, as applied in the previous section, to discussion that is instead prescriptive and summative.
The goal here is to synthesize results from reported annotation-and-reconstruct experiments, as well as other work testing and applying hereditary stratigraphy, to provide concrete, action-oriented advice to guide the reader in effectively applying hereditary stratigraphy in their own digital evolution work.

Discussion covers the following questions,
\begin{enumerate}
\item Should I use hereditary stratigraphy or phylogenetic direct tracking?
\item How should I choose appropriate hereditary stratigraphy configuration for my use case?
\item How do I integrate hereditary stratigraphy instrumentation into my digital evolution simulation?
\item How do I work with hereditary stratigraphy annotation data once it is generated?
\end{enumerate}

\subsection{Hereditary Stratigraphy or Direct Tracking?}

If you are not using parallel or distributed computing, direct phylogenetic tracking should usually be preferred due to its capability for perfect record-keeping; likewise, if your simulation uses a centralized controller-worker paradigm.
A tool like Phylotrack (asexual phylogenies; Python/C++), APoGET (sexual phylogenies; C++) or MABE (C++) might be appropriate for your use case \citep{dolson2024phylotrackpy,bohm2017mabe,godin2019apoget}.
It is also reasonably straightforward to implement phylogeny tracking yourself \citep{moreno2024algorithms}.
However, for serial or centralized simulations, there are a few scenarios where reconstruction-based tracking may be useful,
\begin{itemize}
\item resource-constrained runtime environments where memory is scarce or dynamic memory allocation is not supported (e.g., embedded);
\item simulation objectives require hard real-time operations; or
\item \textit{ad hoc} serialization and re-use of agents across simulation runtimes, where maintaining a cohesive global record is difficult or impractical.
\end{itemize}

In simulations employing decentralized parallel and distributed computation, a reconstruction-based approach such as hereditary stratigraphy is more likely to be appropriate.
Compared to perfect tracking in this setting, hereditary stratigraphy provides simpler implementation, lower runtime communication overhead, and greater robustness to data loss.
Note that if annotation size is not a concern or a low generation count is anticipated, essentially perfect-quality reconstruction can be achieved with hereditary stratigraphy methods.
In this case, one could simply use a retention policy that discards no differentiae and a sufficient differentia width to effectively guarantee collisions will not occur (e.g., 32 or 64 bits).
However, in most cases, a compromise between phylogeny approximation and per-genome annotation size will be necessary.
We cover this in the next section.

\subsection{Annotation Configuration}

The following provides step-by-step instruction on selecting appropriate hereditary stratigraphy methods for a given use case.
We cover
\begin{enumerate}
\item determining annotation order of growth,
\item selecting annotation size,
\item selecting differentia width,
\item picking a retention policy, and
\item selecting between column- or surface-based implementation.
\end{enumerate}

Discussion then turns to a handful of special-case topics.

\subsubsection{Annotation Order of Growth}
\textit{Suggested default choice: constant-size annotation.}

In most scenarios, a constant-size annotation will give better runtime performance and ensure fuller use of available memory resources.
However, if you need hard guarantees on absolute or recency-proportional inference quality, an annotation size that scales $\mathcal{O}(n)$ or $\mathcal{O}(log(n))$, respectively, with generational depth will be necessary.
The \textit{hstrat} Python package provides \texttt{fixed\_resolution\_algo} and  \texttt{recency\_proportional\_resolution\_algo} policies for such cases \citep{moreno2022hstrat}.

\subsubsection{Annotation Size}
\textit{Suggested default choice: 256-bit differentia buffer with 64-bit generation counter.}

Annotation size must compromise between memory-use and communication-bandwidth overhead and quality of reconstructed phylogenies.
In cases where annotation size is not a limiting factor, using 256-bit annotation buffers with single-bit differentiae will discern phylogenetic events with about 13\% recency-relative precision (tilted) or 1\% depth-relative precision (steady) through 1 billion generations.
For full-byte differentiae, discussed below, an annotation size on the order of kilobits would be very robust.
Where memory use is a limiting factor, however, a 64-bit annotation buffer with single-bit differentiae can give good results.

In picking annotation size, some consideration should be given to the phylogenetic scale of experimental use cases.
Triplet distance (accuracy) appears to be largely stable under surveyed increases in population size and number of taxa sampled for reconstruction
However, inner node loss can increase when increasing reconstruction sample size.
Where this is a concern, annotation size may need to be increased commensurate to intended phylogeny tip count.

Current implementations of surface algorithms are limited to buffer sizes that are even powers of two (i.e., 32, 64, 128, etc.).
Where finer gradations are desired, one possibility would be to consider intermediate differentia sizes.
For instance, storing 3-bit differentiae over 32 surface sites would occupy 96 bits.
Alternatively, alternating depositions across a collection of surfaces might be considered (e.g., a 32-bit tilted surface and a 16-bit steady surface).
Note, though, that the latter option would require implementation customizations akin to those used to create the hybrid surface retention policy \citep{moreno2024hsurf}.
Column algorithms are more flexible in buffer sizing, but in the case of tilted retention, they make less full use of available buffer space.

Finally, note that, in addition to differentia values, a generation counter will also need to be stored in genomes.
When working with fixed-width data types, sufficient representational range will be necessary to support the maximum-expected generational depth elapsed in simulation.
For most use cases, a 32- or 64-bit counter value will be appropriate.

\subsubsection{Differentia Retention Policy}
\textit{Suggested default choice: hybrid retention policy.}

Hybrid retention policy is a good choice where mild phylodiversity-enhancing factors (ecology, spatial structure, relaxed selection pressure) are expected, or expectations are unclear.
Across use cases, hybrid retention often achieves reconstruction quality competitive with the better of steady and tilted policies.
Even where it is outperformed by steady or tilted retention, hybrid retention avoids catastrophic failure modes characteristic, in particular, of steady retention.
On platforms where steady policy implementation (necessary as a component of hybrid policy) is not readily available, tilted policy should typically suffice.
If you do not expect strong phylodiversity-enhancing factors (i.e., no ecology, no spatial structure, high selection pressure), tilted policy can reliably outperform hybrid policy.
In rare cases where phylodiversity-enhancing factors are very strong (e.g., pure drift conditions), a steady policy may be appropriate.

\subsubsection{Differentia Width}
\textit{Suggested default choice: use bit-size differentiae.}

Bit-size differentiae maximize the fraction of correct reconstruction outcomes, but can also introduce incorrect reconstruction outcomes.
Byte-size differentiae have very low incorrect reconstruction outcome rates and can reconstruct true polytomies, but have a larger incidence of unresolved reconstruction outcomes (artifactual polytomies).
This trade-off results in bit-size differentiae giving more informative reconstructions, as measured by strict triplet distance error.
If you need very strong guarantees against incorrect reconstruction outcomes, an even larger differentia size (32 or even 64 bits) may be appropriate.

\subsubsection{Column vs. Surface Implementation}
\textit{Suggested default choice: surface implementation.}

If you are using dynamic annotation size, you will need a column-based implementation to allow differentia count growth.
Otherwise, for constant annotation size, surface implementations are much more efficient \citep{moreno2024trackable}.
In the case of tilted retention policy, they also give higher-quality reconstructions.
If using steady policy, column implementation gives higher-quality reconstructions, but surface implementation would be reasonable to gain enhanced runtime performance.
Finally, hybrid policy is currently only provided for surface-based implementation.

Another factor that might influence this decision is the software platform targeted.
Where packages providing hereditary stratigraphy are not available, surface algorithms are easier to implement owing to only needing to implement one update operation: site selection on the fixed-size surface buffer.

\subsubsection{Special-case topic: lineage tags}
\textit{Suggested default choice: not needed in most cases.}

In scenarios where explicitly differentiating between founding clades is paramount, consider adding a systematically assigned founder ID or randomly generated fixed tag.
This tag would then be used as a first pass to divvy up end-state annotations into independent sets before feeding them into separate reconstruction processes.
When using single-bit differentiae, if several founding lineages persist, spurious collisions can make completely independent clades falsely appear to share some common history.

\subsubsection{Special-case topic: incorporating trait data}
\textit{Suggested default choice: collect ``fossil'' phenotypes instead.}

It is possible to ``annotate'' differentiae with information about genetic or phenotypic traits associated with corresponding ancestors.
Conveniently, because every internal node in a hereditary stratigraphy reconstruction corresponds to a retained differentia, this approach ensures all internal nodes in a reconstructed phylogeny can be associated with a trait value.
However, a similar result can be had by saving out sample ``fossil'' specimens (with corresponding trait information) over the course of a simulation.
Because, like extant taxa, these fossils have hereditary stratigraphy annotations, they can readily be incorporated into phylogenetic reconstruction and provide information on ancestral trait states.
This approach avoids bloating annotation sizes with dozens of trait values on each genome.

\subsection{Runtime Integration}

Two major steps are necessary to integrate hereditary stratigraphy into evolution simulations: (1) add instrumentation to the \texttt{Genome} data type and (2) hook annotation update procedures to the copy/reproduce or mutate routine.

Software implementing column-based approaches is available for Python \citep{moreno2022hstrat} and surface-based approaches are available for Python \citep{moreno2024hsurf} and Zig/Cerebras Software Language \citep{moreno2024wse}.
Based on community feedback, C/C++ and Rust are priority targets for surface-based implementation ports.
However, porting core surface algorithms to another software language should be doable with moderate effort.
Surface algorithms are succinct (fewer than 30 lines of code) and accompanying unit tests are provided, making translation to new programming languages straightforward \citep{moreno2024algorithms}.
We found LLM assistance to be highly effective in translating the surface algorithms from Python to Zig.
That said, we encourage interested researchers to reach out if hereditary stratigraphy implementation is not available in their chosen programming language --- we would be happy to collaborate in providing implementations in additional languages where there is a use case.

\subsubsection{Annotation Data: Asexual}

For surface-based annotations, two data components are required: a fixed-size differentia buffer and a generation counter.
The differentia store can be implemented as an array of integer data types (e.g., \texttt{uint8}), but for bit differentiae you will likely want to use a raw memory array or an abstraction around it if available (e.g., \texttt{std::bitset}).
Differentia store memory should be randomized upon initialization.

For column-based annotations, the \textit{hstrat} Python library furnishes a prepackaged \texttt{HereditaryStratigraphicColumn} class, which encapsulates all necessary state.

\subsubsection{Annotation Data: Sexual}
\textit{Suggested default choice: for sexual phylogenies, inherit annotation from maternal (or arbitrary) parent, akin to mtDNA.}

The core of existing hereditary stratigraphy is designed around asexual lineages.
Some preliminary work has demonstrated applications of hereditary stratigraphy to sexual populations \citep{moreno2024methods}, but canonical protocols have yet to be established.
One possibility includes tracking asexual lineages of individual genes (``gene trees'').
In this scenario, annotation data would be associated to individual genetic building blocks.
Organism-level annotation could be used to track the emergence of new independently-breeding subpopulations (``species trees''').
This approach relies on distinct annotation values reaching fixation within independent subpopulations.
Fixation can be left to drift (where offspring inherit the differentia record of an arbitrary parent).
Alternatively, a gene drive mechanism can be applied.
Such a mechanism favors inheritance of large-magnitude differentia values, thereby causing them to sweep through interbreeding subpopulations.
Refer to \citet{moreno2024methods} for more detailed discussion of this topic.

\subsubsection{Annotation Update}
\textit{Suggested default choice: update hereditary stratigraphy annotations every generation.}

When annotations are inherited, they should be updated to add a new differentia and increment the annotation-associated generation counter.
Usually, this should be accomplished by adding logic into an existing create offspring routine.
For genotype-level tracking (as opposed to individual-level tracking), annotation updates could instead be conducted as a part of an ``apply mutation'' routine.

When using the \textit{hstrat} \texttt{HereditaryStratigraphicColumn} object, update logic is bundled into the \texttt{DepositStratum} method.

For surface-based annotations, call the chosen retention policy's \texttt{pick\_deposition\_site} method and randomize the differentia element at the returned $n$th position.
Then, the associated generation counter should be incremented.

For most applications, annotation updates (i.e., appending a new differentia) should correspond directly to generations elapsed.
However, in cases where fine-resolution visibility below a certain level is not useful, annotations may be updated less frequently.
For tilted policy, in particular, this approach can help preserve ancient differentia coverage for longer.

In circumstances where time-indexed resolution or ultrametricity (tips sharing equal branch lengths from root) is necessary, annotations can be updated on the basis of logical simulation time or even real-time rather than generations elapsed.
In this case, annotation updates would be applied, as necessary, on parents' annotations to bring them up-to-date with the current simulation time whenever a reproduction event occurs.
Such an approach may require multiple back-to-back updates if several simulation time clock cycles have elapsed.
An alternate approach to achieve a time-calibrated tree would be to attach simulation timestamps to differentiae or to record time-stamped ``fossil'' taxa, as discussed above.

\subsection{Postprocessing and Analysis}

\subsubsection{Sampling strategies}

A typical approach will sample end-state extant agents as taxa for phylogenetic reconstruction.
Note that it is not necessary to exhaustively collect the entire end-state population; for many analysis use cases, a representative sample of population members will suffice.
Such an approach, however, will not provide information on extinct lineages.
Because hereditary stratigraphy reconstruction can stitch together taxa from across widely varying time points, specimens can be sampled during an evolutionary run and be integrated directly into phylogenetic reconstruction with end-state extant genomes.
Such intermediate taxa from earlier time points can serve a role akin to ``fossils'' in natural history.
For more advanced analyses, it may be desirable to associate sampled specimens with genetic and/or phenotypic trait data \citep{dolson2019modes,khabbazian2016fast}.

\subsubsection{Annotation Serialization}

To export annotation data for analysis, you will need to save it to file.
Workflows that save annotations separately from other genome components using a conventional plain text format (e.g., JSON, CSV) are typically most convenient.
The \textit{hstrat} library provides a number of plug-and-play utilities for serialization of \texttt{HereditaryStratigraphicColumn} objects.
Functionality to provide direct support for serializing surface data is a priority item on the \textit{hstrat} road map.
In the meantime, this must be done manually.
For each annotation, you will need to store the generation counter and the differentia data.
Hex strings provide one potentially convenient approach to encode differentia data.
When loading annotation data, you will need to know the retention policy and differentia width used --- so make sure these are recorded.

Note that while the \textit{hstrat} library also provides support to serialize/deserialize from compact binary formats, storage in plain text format with zipping (e.g., gzip) will often provide equivalent space efficiency to binary representations and can be considerably easier to work with.

Phylogenetic reconstruction is implemented in the \textit{hstrat} Python library, so it will typically be necessary to load serialized annotation data into this environment.
Functions to convert raw data into \texttt{Column} object instances are described in package documentation and examples, including utilities compatible with JSON, YAML, and CSV formats.
Functionality to provide direct support for deserializing surface data is also a priority item on the \textit{hstrat} road map.
In the meantime, this functionality is provided separately \citep{moreno2024hsurf}.

\subsubsection{Phylogeny Reconstruction}

Phylogenetic reconstruction from annotations is implemented as \textit{build\_tree} in the \textit{hstrat} Python library \citep{moreno2022hstrat}.
This algorithm takes in a sequence of deserialized column objects and produces a phylogeny.
Optionally, if desired, a list of taxon identifiers can be provided to label leaf nodes in the reconstructed phylogeny.

Utilities to directly estimate MRCA generation between column objects without performing full reconstruction are also available in \textit{hstrat}.

\subsubsection{Phylogeny Analysis}

Phylogenetic reconstructions from \textit{hstrat} are returned as a Pandas \texttt{DataFrame} in alife community data standard format \citep{lalejini2019data,reback2020pandas}.
Tools are available to convert alife standard data into standard bioinformatics formats, such as Newick, NeXML, and NEXUS \citep{moreno2024apc}.
This interconvertibility allows interoperation with rich existing software ecosystems for phylogenetic visualization and analysis.
The online tool IcyTree provides a good starting point; phylogeny data can be uploaded in Newick format to create rich, exportable tree visualizations fully in-browser \citep{vaughan2017icytree}.

\section{Conclusion} \label{sec:conclusion}

In this work, we have applied empirical annotate-and-reconstruct experiments to benchmark inference quality of hereditary stratigraphy approaches across use cases varying in phylogenetic structure, scale, and allocated annotation space.
In these experiments, we consider,
\begin{itemize}
\item \textbf{differentia retention:} whether annotation space should be allocated for finer resolution in discerning recent phylogenetic events,
\item \textbf{annotation implementation:} comparing existing column-based approaches to newer surface-based approaches optimized for fixed-size annotations, and
\item \textbf{differentia width:} how many bits should be used per lineage checkpoint to reduce the probability of spuriously overestimating relatedness.
\end{itemize}

Findings are then applied to develop practitioner-oriented guidelines to effectively employ hereditary stratigraphy methodology.
Principal results are,
\begin{enumerate}
\item tilted retention produces better reconstruction quality than steady retention, except in scenarios with very high phylogenetic richness (e.g., drift conditions);
\item hybrid tilted-steady retention provides good reconstruction quality across scenarios;
\item for tilted retention, surface-based implementation provides better reconstruction quality than column-based implementation;
\item for steady retention, column-based implementation provides better reconstruction quality than surface-based implementation; and
\item increased differentia size increases accuracy but reduces precision.
\end{enumerate}
As tilted policy is likely to be preferred in practice, it is promising to see surface-based implementation improve reconstruction quality in this case.
Because surface-based approaches were designed foremost to optimize performance and be easier to code for new platforms (particularly in low-level environments) \citep{moreno2024trackable}, additionally achieving enhanced reconstruction quality makes their adoption a win-win situation.

Owing to its inspiration from inference-based phylogenetics work in biology, hereditary stratigraphy is designed to operate in an entirely decentralized manner that is, by nature, efficient to scale and robust to disruptions or data loss.
It is therefore promising to see that reconstruction accuracy of hereditary stratigraphy is also generally robust to scale-up.
On the other hand, we found inner node loss --- a precision measure --- to be sensitive to increases in the number of taxa sampled for reconstruction.
This issue, with bit-width differentiae, arises due to increased probability for exactly identical annotations through differentia collision, resulting in clumping of tip nodes into polytomies.
This problem may be abated in systems with nonsynchronous generations, where tips are spread apart by generational depth.
That said, we did find inner node loss to be largely robust to scale-up of the actual population size of a simulation, with the number of taxa sampled for reconstruction held constant.

Our goal in developing hereditary stratigraphy is to provide methodology that is sufficiently lightweight, modular, and flexible for general-purpose use across digital evolution systems.
Here, we have provided a comprehensive, evidence-driven foundation for effective application of hereditary stratigraphy across experimental use cases.
Explicitly compiling this material as a prescriptive guide maximizes its utility to this end.
However, we anticipate that --- most of all --- adoption hinges on success in providing a seamless, plug-and-play developer experience to those wishing to incorporate the methodology.
As such, we seek to provide packaged library software with easy-to-learn API design and thorough documentation.
Note that, beyond content presented here, the \textit{hstrat} repository includes a small library of code samples demonstrating end-to-end use of hereditary stratigraphy, useful as a starting point for new users \citep{moreno2022hstrat}.
We would be very interested in collaborating to integrate hereditary stratigraphy instrumentation into your system or to develop algorithm implementations for your particular programming language and runtime environment.

Present work motivates several further steps in developing hereditary stratigraphy methodology.
From a practical perspective, we wish to make improvements in curating public-facing surface-based implementations that are interoperable with existing column-based tools.
Another practical consideration will be optimization, and perhaps parallelization, of reconstruction to support work with very large taxon sets.
In a separate vein, accuracy loss from differentia collisions when working with bit-level differentia may warrant effort in developing means to sample among possible collision sets and generate a consensus tree with accompanying uncertainty measures \citep{bryant2003classification}.

Considering a broader perspective on future work, development of hereditary stratigraphy comprises only one aspect of a broader agenda in scaling up digital evolution experiments.
Among other avenues, research will need to explore simulation synchronization schemes \citep{fujimoto1990parallel}, best-effort computing approaches \citep{moreno2022best,ackley2020best}, emerging hardware architectures \citep{moreno2024trackable,chan2018lenia,heinemann2008artificial}, and scalable assays for evolutionary innovation, ecological dynamics, and various forms of complexity \citep{bedau1998classification,dolson2019modes,moreno2024methods,moreno2024case,moreno2024ecology}.

\section*{Acknowledgement}

This research was supported by Michigan State University through the computational resources provided by the Institute for Cyber-Enabled Research and is based upon work supported by the Eric and Wendy Schmidt AI in Science Postdoctoral Fellowship, a Schmidt Sciences program.
This research was supported in part by funding from the NSF (DEB 1813069).
Any opinions, findings, and conclusions or recommendations expressed in this material are those of the author(s) and do not necessarily reflect the views of the National Science Foundation.

\putbib

\end{bibunit}

\clearpage
\newpage

\begin{bibunit}

\section{Supplemental Material}

\begin{figure*}
  \centering
  \includegraphics[width=0.9\textwidth]{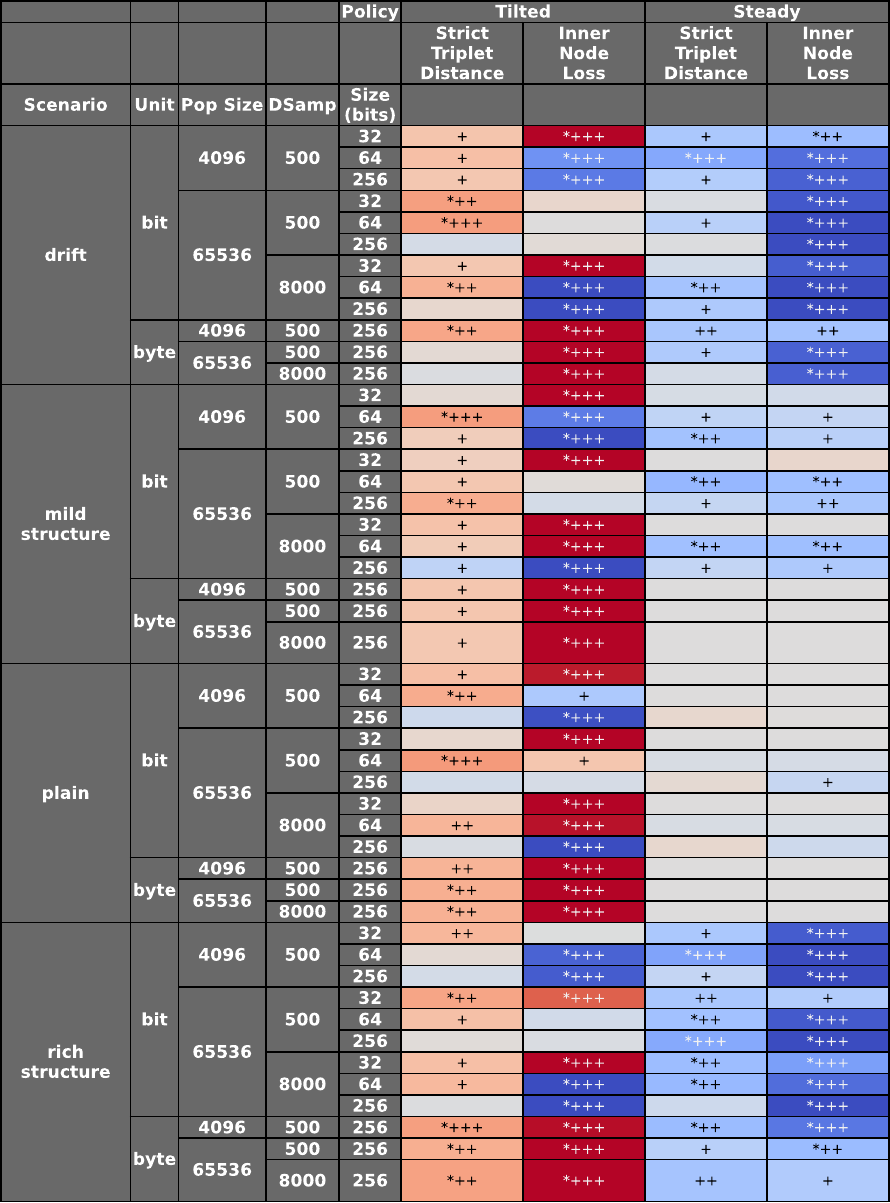}
  \caption{%
   \textbf{Comparison of reconstruction quality from column- and surface-based annotations.}
   \footnotesize
    Color coding reflects non-parametric comparison between quality measure values, with red indicating superior surface performance and blue indicating superior column performance.
    Left column shows tilted retention policies, and right column shows steady retention policies.
    In cell annotations, +'s indicate small, medium, and large effect sizes using the Cliff's delta statistic and *'s indicate statistical significance at $\alpha = 0.05$ via Mann-Whitney U test.
  }
  \label{fig:col-vs-surf}
\end{figure*}

\end{bibunit}

\end{document}